\documentclass{article}

\PassOptionsToPackage{table}{xcolor}
\usepackage{iclr2027_conference,times}

\usepackage{graphicx}
\usepackage{booktabs}
\usepackage{xcolor}
\usepackage[hidelinks]{hyperref}
\usepackage{url}
\usepackage{etoolbox}
\usepackage{orcidlink}

\usepackage{algorithm}
\usepackage{algorithmicx}
\usepackage{algpseudocode}
\usepackage{float}
\usepackage{dblfloatfix}
\usepackage{array}
\usepackage{multirow}
\usepackage{wrapfig}
\restylefloat{algorithm}
\usepackage{makecell}
\usepackage{siunitx}
\usepackage{subcaption}
\usepackage{tikz}
\usetikzlibrary{arrows.meta}

\usepackage[normalem]{ulem}
\usepackage{pifont}

\usepackage{amsmath,amsfonts,bm}

\def\eqref#1{equation~\ref{#1}}

\def\1{\bm{1}}

\DeclareMathAlphabet{\mathsfit}{\encodingdefault}{\sfdefault}{m}{sl}
\SetMathAlphabet{\mathsfit}{bold}{\encodingdefault}{\sfdefault}{bx}{n}

\usepackage{xspace}

\newcommand{\DDTs}{\DDTs\xspace}

\definecolor{customgreen}{HTML}{E6F8E0}

\definecolor{ggreen}{rgb}{0.53, 0.69, 0.43}
\definecolor{gred}{HTML}{F5433D}

\definecolor{deemph}{gray}{0.6}

\newcommand{\gc}[1]{\textcolor{deemph}{#1}}

\makeatletter
\patchcmd{\@maketitle}
  {\begin{tabular}[t]{l}\bf\rule{\z@}{24pt}\@author\end{tabular}}
  {\noindent\hbox to \textwidth{
    \begin{tabular}[t]{@{}l@{}}\bf\rule{\z@}{24pt}\@author\end{tabular}\hfil}}
  {}{}
\makeatother

\title{RAC: Rectified Flow Auto Coder}

\author{%
\textbf{Sen Fang} \hspace{0.5em}
\textbf{Yalin Feng}\textsuperscript{*} \hspace{0.5em}
\textbf{Yanxin Zhang} \hspace{0.5em}
\textbf{Yihao Quan} \hspace{0.5em}
\textbf{Juyi Lin} \hspace{0.5em}
\textbf{Yifan Shen} \\
\textbf{Ziwei Dong} \hspace{0.5em}
\textbf{Sisong Bei} \hspace{0.5em}
\textbf{Dimitris N. Metaxas}\textsuperscript{\textdagger}
\\[0.5em]
\textsuperscript{*}Equal contribution \quad \textsuperscript{\textdagger}Corresponding author
\\[0.5em]
\textbf{Project Page:} \href{https://world-snapshot.github.io/RAC}{\textcolor{magenta}{\textbf{\nolinkurl{https://world-snapshot.github.io/RAC}}}}
}

\iclrfinalcopy

\begin{document}

\maketitle
\lhead{Preprint}

\vspace{-32pt}
\begin{figure}[h]
  \centering
  \includegraphics[width=\linewidth]{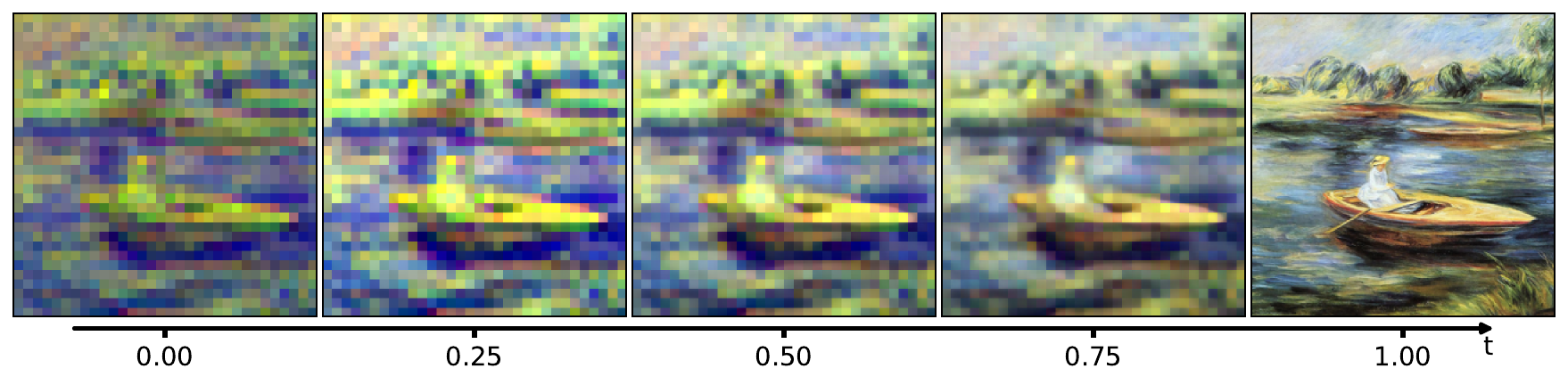}
  \caption{\textbf{The trajectory demonstration of RAC:} Make the reconstruction task a condition generation task; Make the decoder the encoder; Make the single-step decoding and encoding a multi-step decoding and encoding.}
  \label{fig:rac-teaser}
\end{figure}

\begin{abstract}

In this paper, we propose a \textbf{R}ectified Flow \textbf{A}uto \textbf{C}oder (\textbf{RAC}) inspired by Rectified Flow to replace the traditional VAE:
\textbf{\textcolor{purple}{1.}} It achieves multi-step decoding by applying the decoder to flow timesteps. Its decoding path is straight and correctable, enabling step-by-step refinement.
\textbf{\textcolor{purple}{2.}} The model inherently supports bidirectional inference, where the decoder serves as the encoder through time reversal (hence \textit{Coder} rather than encoder or decoder), reducing parameter count by nearly 41\%.
\textbf{\textcolor{purple}{3.}} This generative decoding method improves generation quality since the model can correct latent variables along the path, partially addressing the reconstruction--generation gap.
Experiments show that RAC achieves a Pareto improvement over SOTA VAEs, where even a 10× parameter-reduced decoder exceeds full-scale VAE performance in both reconstruction and generation quality, validating the effectiveness of our approach.

\end{abstract}

\section{Introduction}
\label{sec:intro}

In the field of generative models, over the past many years, there has been a problem of inconsistent generation and reconstruction results. For instance, when using the same generation method and the same VAE, the generation results are often inferior to the reconstruction results: \textbf{\textcolor{purple}{(1)}} Some studies consider it to be a problem of representation \cite{zheng2026diffusion,wang2026vaerepavariationalautoencoderrepresentation}, and attempt to solve it by optimizing the latent space; \textbf{\textcolor{purple}{(2)}} Some studies consider it to be a problem of training strategy \cite{leng2025repaeunlockingvaeendtoend}, and use the generation backbone and VAE for joint training optimization; \textbf{\textcolor{purple}{(3)}} Some studies consider it to be a problem of the method \cite{zhuang2025vargptunifiedunderstandinggeneration}, and propose a native unified method to try to solve it, etc.

Under the same conditions of using VAE \cite{kingma2014auto}, one very obvious and unique variable that distinguishes generation from reconstruction is - the variables used during generation are provided by the generation framework (such as Unet, DiT, etc.). Therefore, we hypothesize that part of the reason is that the latent variables output by the generation method are problematic. \textbf{It easily misses the uneven manifold space of traditional VAE}, resulting in unstable generation effects and being less accurate than the encoder provided by VAE itself\footnote{From another perspective, Text2Img (prediction, information increase) is not as precise as Img2Latent (encoding compression, information reduction), which can also serve as a reason for the disparity in the generation and reconstruction capabilities.}.

So, in the case of minimal modifications, the method to enhance the decoder's performance during generation is to incorporate the decoder into the generation process. As shown in Fig. \ref{fig:rac-teaser}, we introduce the concepts of time steps and velocity fields into the decoder, \textbf{enabling it to gradually calibrate the input variables or the behavior during encoding}, and theoretically, it inherently supports bidirectional transformation. The decoder of traditional VAE is an important part of the generation process. Generation frameworks (such as Unet \cite{10.1007/978-3-319-24574-4_28}, DiT \cite{10377858}, etc.) can perform multi-step reasoning, but the end VAE is actually a one-step generation. This is actually not very reasonable: Consider a navigator who, despite having a correctable route, is forced to teleport directly to the destination in one step — losing all opportunity to adjust course along the way.

Specifically, we take a unified view of continuous-time generation and representation learning, and elevate the classic VAE decoder from a pointwise mapping to a \emph{path-level} generative process: a time-conditioned velocity field defines an integrable flow from latent variables to the image space, and explicitly models a correctable, multi-step decoding trajectory. This design provides a structured generation path while making the same model naturally act as an encoder under time reversal, enabling parameter sharing and bidirectional consistency. For training, we center on path consistency, latent alignment, and reconstruction constraints, yielding a unified optimization framework that balances invertibility and generation quality. Our contributions are threefold:

  

\begin{itemize}
    \item We propose RAC, which generalizes VAE decoding from a \emph{single-step map} to a \emph{continuous-time, integrable path}, establishing a \textbf{unified flow-based autoencoding paradigm} for generation and representation.
    \item We design a \textbf{structured bidirectional mechanism} where the same velocity-field model performs decoding and encoding in forward and reverse time, achieving parameter sharing, path reversibility, and generation--reconstruction consistency.
    \item We present a stable training objective and implementation recipe (\textbf{path consistency + latent alignment + reconstruction constraints}), delivering strong reconstruction and generation performance at comparable model scale.
\end{itemize}

\section{Methodology}
\label{sec:method}

\begin{figure}[t]
\centering
\includegraphics[width=0.95\linewidth]{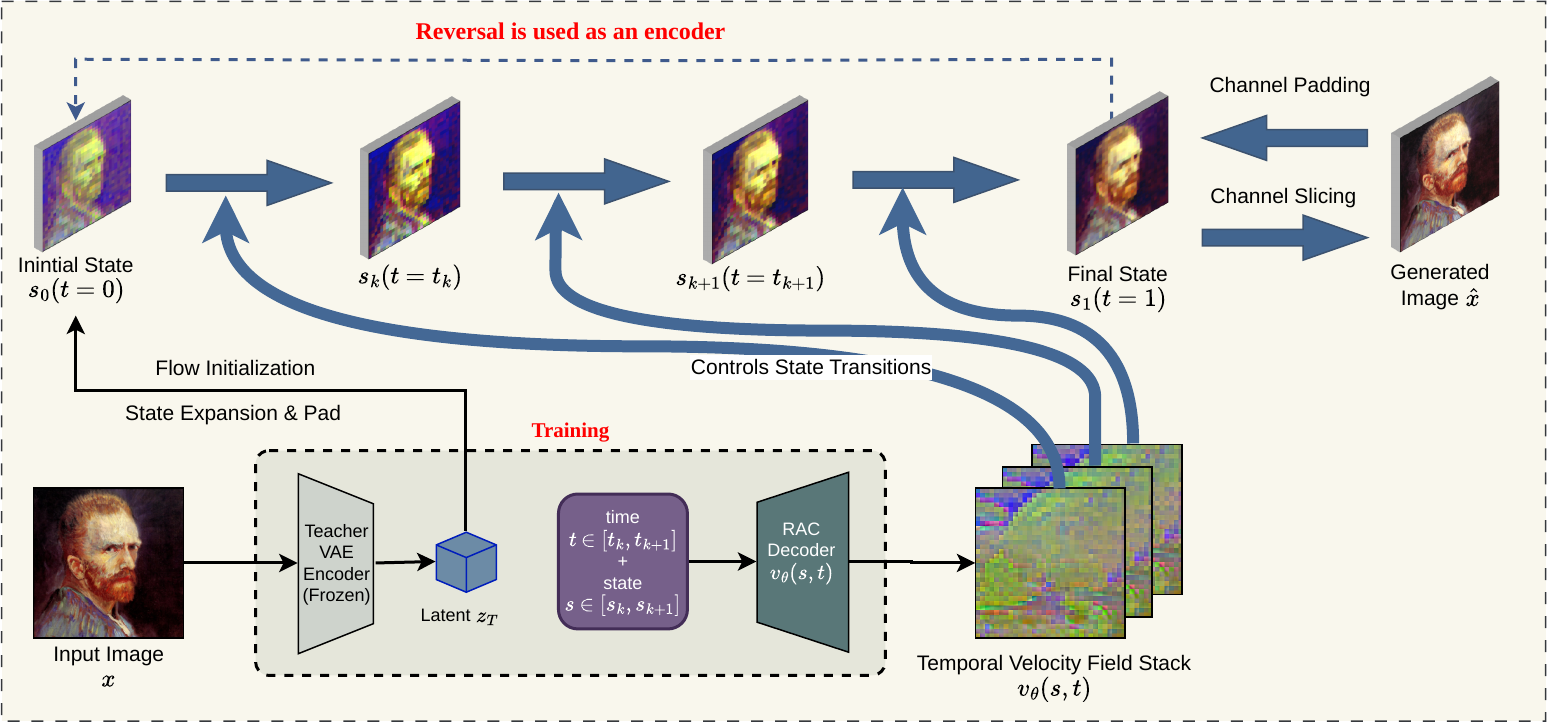}
\caption{\textbf{Method Overview.}
\textbf{\textcolor{purple}{(i) Training.}} To prevent latent space collapse, we freeze the VAE encoder and train only the RAC decoder; reverse-time inference then serves as encoding.
\textbf{\textcolor{purple}{(ii) State Construction.}} Extra channels beyond RGB are padded with 0.5, keeping the velocity field shape constant and ensuring bidirectional consistency.
\textbf{\textcolor{purple}{(iii) RAC Input.}} RAC takes time $t$ and the current state as input, driving the transition from latent initialization to the target image.
}
\label{fig:rac-method}
\vspace{-12pt}
\end{figure}

Figure~\ref{fig:rac-method} illustrates the RAC pipeline.
We replace the single-step VAE decoder with a continuous-time velocity field $\mathbf{v}_\theta$, which integrates a full-resolution \emph{state} tensor $\mathbf{s} \in \mathbb{R}^{C_s \times H \times W}$ from a latent-derived initialization to the target image; time-reversal of the same model performs encoding.
To prevent latent space collapse, we freeze the VAE encoder and train only the decoder under a joint objective covering reconstruction, path consistency, and latent alignment.

\subsection{Overview and Goal}
\label{sec:method_overview}

The core idea is to replace the single-step VAE decoder with a \emph{time-conditioned velocity field} $\mathbf{v}_\theta(\mathbf{s}, t)$ that integrates a \emph{state} $\mathbf{s}$ from a latent-derived initialization $\mathbf{s}_0$ to a target image state $\mathbf{s}^*$; running the same field in reverse serves as encoding.
This formulation targets the fundamental mismatch in VAE-style models: reconstruction uses a data-conditioned encoder, while generation relies on latents that may not respect the decoder's learned manifold. By making decoding a multi-step, correctable process, generation can actively refine latent variables along the path rather than committing to a one-shot projection.


Let $\mathbf{x}\in[0,1]^{3\times H\times W}$ be an image and $\tilde{\mathbf{x}}=2\mathbf{x}-1\in[-1,1]^{3\times H\times W}$ its normalized version. We define a state tensor by padding channels:
\begin{equation}
\mathbf{s}=\mathrm{pad}(\tilde{\mathbf{x}})\in\mathbb{R}^{C_s\times H\times W}, \quad C_s\ge 3.
\end{equation}
\begin{wrapfigure}{R}{0.40\textwidth}
  \vspace{-22pt}
  \centering
  \includegraphics[width=\linewidth]{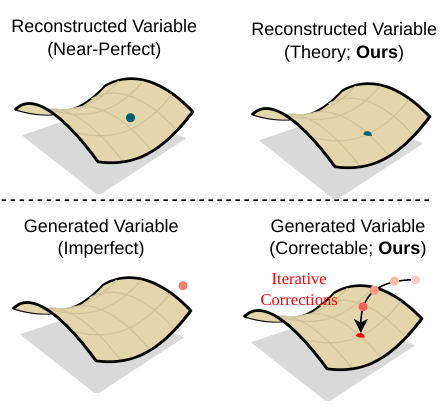}
  \vspace{-20pt}
  \caption{\textbf{Reconstruction is Condition Generation:} The previous reconstructions were more accurate because they could relatively approach the manifold. The previous generations predicted variables that were often some distance away from the manifold. This is part of the reason for the past differences in the performance of generation and reconstruction. However, our method theoretically aims for perfect reconstruction, and the multi-step decoding can correct the potential variables provided by Unet/DiT. Therefore, both reconstruction and generation are significantly superior to traditional VAEs.}
  \label{fig:4figs}
  \vspace{16pt}
\end{wrapfigure}

A KL-VAE teacher provides a latent target $\mathbf{z}_T\in\mathbb{R}^{C\times h\times w}$. We learn a continuous-time flow from latent state to image state and reuse the same flow in reverse for encoding. This gives us a single model that \emph{(i)} decodes with step-wise correction, \emph{(ii)} supports bidirectional inference, and \emph{(iii)} aligns generation and reconstruction within one mechanism. This formulation is intentionally generic: any KL-regularized VAE can serve as the teacher, which makes the method broadly compatible with existing pipelines.

\subsection{Time-Conditioned Rectified Flow Decoder}
\label{sec:method_flow}
We model the decoder as a time-conditioned velocity field:
\begin{equation}
\frac{d\mathbf{s}(t)}{dt} = \mathbf{v}_\theta(\mathbf{s}(t), t), \quad t\in[0,1].
\end{equation}
The velocity field is predicted by a lightweight VAE backbone \cite{Rombach_2022_CVPR}. To stabilize and structure the flow, the network uses the current state, an explicit time channel, and optional relative positional encoding to return a full-resolution velocity field:
$\mathbf{v}_\theta(\mathbf{s},t) = f_\theta\Big(\mathrm{down}(\mathbf{s}), \; t\mathbf{1}, \; \mathrm{pos}(\mathbf{s})\Big)\in\mathbb{R}^{C_s\times H\times W}$.

We build a latent state by padding the teacher latent and spatially expanding it:
\begin{equation}
\mathbf{s}_0 = \mathrm{expand}\big(\mathrm{pad}(\mathbf{z}_T)\big)\in\mathbb{R}^{C_s\times H\times W}.
\end{equation}
Decoding is then performed by integrating the flow from $t=0$ to $t=1$:
\begin{equation}
\mathbf{s}_K = \mathbf{s}_0 + \int_0^1 \mathbf{v}_\theta(\mathbf{s}(t), t)\, dt,
\end{equation}
and the predicted RGB image is the projection $\hat{\mathbf{x}}=\Pi(\mathbf{s}_K)$ onto the first three channels. Crucially, the \emph{same} model yields an encoder by reversing time:
\begin{equation}
\mathbf{s}_0 = \mathbf{s}_K - \int_0^1 \mathbf{v}_\theta(\mathbf{s}(t), t)\, dt.
\end{equation}
In practice we use $K$ Euler steps with optional random time grids to encourage robustness:
\begin{equation}
\mathbf{s}_{k+1} = \mathbf{s}_k + \Delta t_k\, \mathbf{v}_\theta(\mathbf{s}_k, t_k), \quad k=0,\ldots,K-1,
\end{equation}
and inject small Gaussian noise per step to prevent brittle dynamics. This step-wise integration explicitly exposes intermediate states, enabling both diagnostic visualization and targeted regularization of the decoding path.

As shown in Figure~\ref{fig:4figs}, RAC first learns image-state reconstruction, then organizes the latent space through generation, and finally refines decoding through multi-step calibration. This training objective directly bridges generation and reconstruction.

\begin{figure}[t]
\centering
\includegraphics[width=\linewidth]{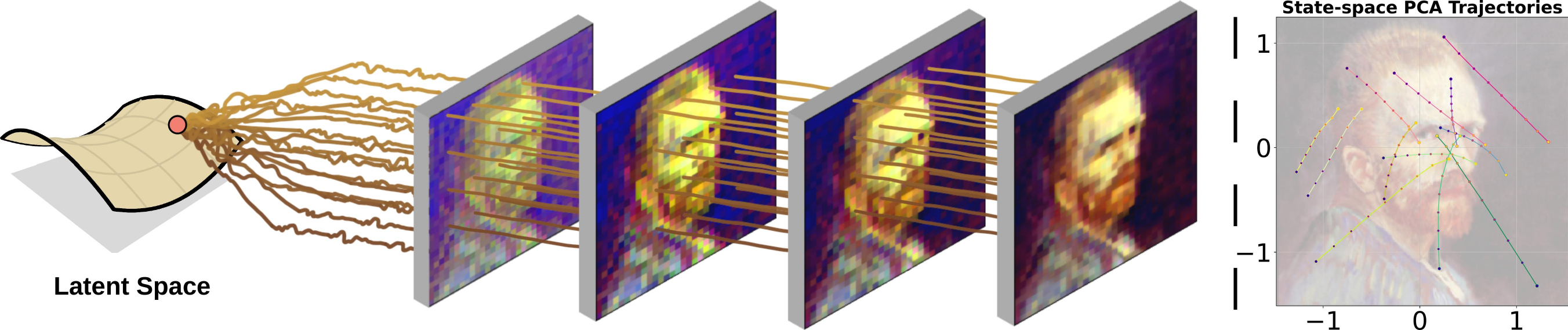}
\vspace{-12pt}
\caption{\textbf{Conceptual and empirical views of RAC generation trajectories.} Left: a conceptual illustration of trajectory-based generation from latent space to image space. Right: sampled state trajectories projected into a 2D PCA space.}
\label{fig:rac-path}
\vspace{-12pt}
\end{figure}

\subsection{State Construction and Transition}
\label{sec:method_state}
To bridge the resolution gap between latent space and image space, we introduce explicit \emph{state construction} operators. The KL-VAE latent is padded and spatially expanded into a full-resolution latent state, $\mathbf{s}_0 = \mathrm{expand}\big(\mathrm{pad}(\mathbf{z}_T)\big)\in\mathbb{R}^{C_s\times H\times W}$.
The target image state is defined by padding the normalized image:
$\mathbf{s}^\ast = \mathrm{pad}(\tilde{\mathbf{x}})$.
The transition $\mathbf{s}_0\rightarrow\mathbf{s}^\ast$ is performed by integrating the rectified flow, producing a trajectory $\{\mathbf{s}_k\}_{k=0}^{K}$:
\begin{equation}
\mathbf{s}_{k+1} = \mathbf{s}_k + \Delta t_k\, \mathbf{v}_\theta(\mathbf{s}_k, t_k), \quad k=0,\ldots,K-1.
\end{equation}
The inverse transition is obtained by time reversal using the same velocity field, enabling a single-model encoder--decoder pair. This explicit state transition view makes the generation path interpretable and allows step-wise correction in both directions, which is particularly important when the generated latent deviates from the encoder manifold. The encoded full-resolution state is downsampled back to the teacher latent size via a lightweight learned projector or average pooling. When a teacher latent channel count differs from $C_s$, a channel projection maps it to the RAC state width before spatial expansion.

As shown in Fig.~\ref{fig:rac-path}, RAC generation can be viewed as a trajectory-based process rather than a single-step mapping. The left panel provides an intuitive illustration of this perspective, where a latent representation evolves through a sequence of intermediate image states. The right panel offers an empirical counterpart by projecting sampled state trajectories into a 2D PCA space. These projected paths suggest that RAC follows structured and coordinated routes in its shared state space during generation. For more engineering details, please refer to Section \ref{sec:details}.

\subsection{Training Objectives}
\label{sec:method_objectives}
Our losses are designed to enforce three properties: \emph{accurate reconstruction, rectified paths, and latent consistency}. Let $\mathbf{s}^\ast=\mathrm{pad}(\tilde{\mathbf{x}})$ be the target image state and $\{\mathbf{s}_k\}_{k=0}^{K}$ the decoded trajectory. All squared norms below denote mean squared error averaged over batch and tensor elements. We define a standard reconstruction loss:
$\mathcal{L}_{\text{recon}} = \big\|\mathbf{s}_K - \mathbf{s}^\ast\big\|_2^2$.
To explicitly encourage a \emph{uniform, correctable} path, we penalize deviation from linear interpolation:
\begin{equation}
\vspace{-4pt}
\mathcal{L}_{\text{path}}^{\text{dec}}
= \frac{1}{K-1}\sum_{k=1}^{K-1}
\Big\|\mathbf{s}_k-\mathbf{s}_0-t_k(\mathbf{s}^\ast-\mathbf{s}_0)\Big\|_2^2.
\vspace{-2pt}
\end{equation}

For encoding, we reverse the flow to obtain $\hat{\mathbf{s}}_0$, downsample it to latent $\hat{\mathbf{z}}$, and align it with the teacher latent: $\mathcal{L}_{\text{latent}} = \|\hat{\mathbf{z}} - \mathbf{z}_T\|_2^2$.
We further require that the teacher decoder applied to $\hat{\mathbf{z}}$ matches the input image:
$\mathcal{L}_{\text{pixel}} = \|\mathrm{Dec}_T(\hat{\mathbf{z}}) - \mathbf{x}\|_2^2$.
Finally, we enforce round-trip consistency so that encoding followed by decoding returns to the same image state: $\mathcal{L}_{\text{rt}} = \big\|\mathrm{Flow}_\theta(\mathrm{Flow}^{-1}_\theta(\mathbf{s}^\ast)) - \mathbf{s}^\ast\big\|_2^2$.

The final objective is:
\begin{equation}
\label{eq:rac_objective}
\mathcal{L} = \mathcal{L}_{\text{recon}}
+ \lambda_{\text{path}}\mathcal{L}_{\text{path}}^{\text{dec}}
+ \lambda_{\text{latent}}\mathcal{L}_{\text{latent}}
+ \lambda_{\text{pixel}}\mathcal{L}_{\text{pixel}}
+ \lambda_{\text{rt}}\mathcal{L}_{\text{rt}}
\end{equation}
These objectives jointly enforce accurate endpoints, rectified intermediate paths, and latent alignment, which is crucial for maintaining consistency between reconstruction and generation.

\section{Experiments}

In this section, we present a comprehensive evaluation of RAC, including reconstruction quality comparison (\S\ref{sub:Reconstruction_Quality}), generation quality comparison (\S\ref{sec:gen}), bidirectional consistency and parameter efficiency (\S\ref{sec:bidir}), generation--reconstruction gap analysis (\S\ref{sec:gap}), ablation study (\S\ref{sec:ablation}), and visualization evaluation (\S\ref{sec:visual}).

\vspace{-12pt}
\paragraph{Setup.} All the datasets used in our experiments are the same as those of the comparison models. The detailed experimental settings will be explained in the corresponding sections.

\subsection{Reconstruction Quality}
\label{sub:Reconstruction_Quality}

\begin{figure*}[t]
\vspace{-12pt}
\centering
\hfill
    \raisebox{3pt}{%
    \begin{subfigure}[t]{0.48\textwidth}
        \centering
        \includegraphics[width=0.83\linewidth]{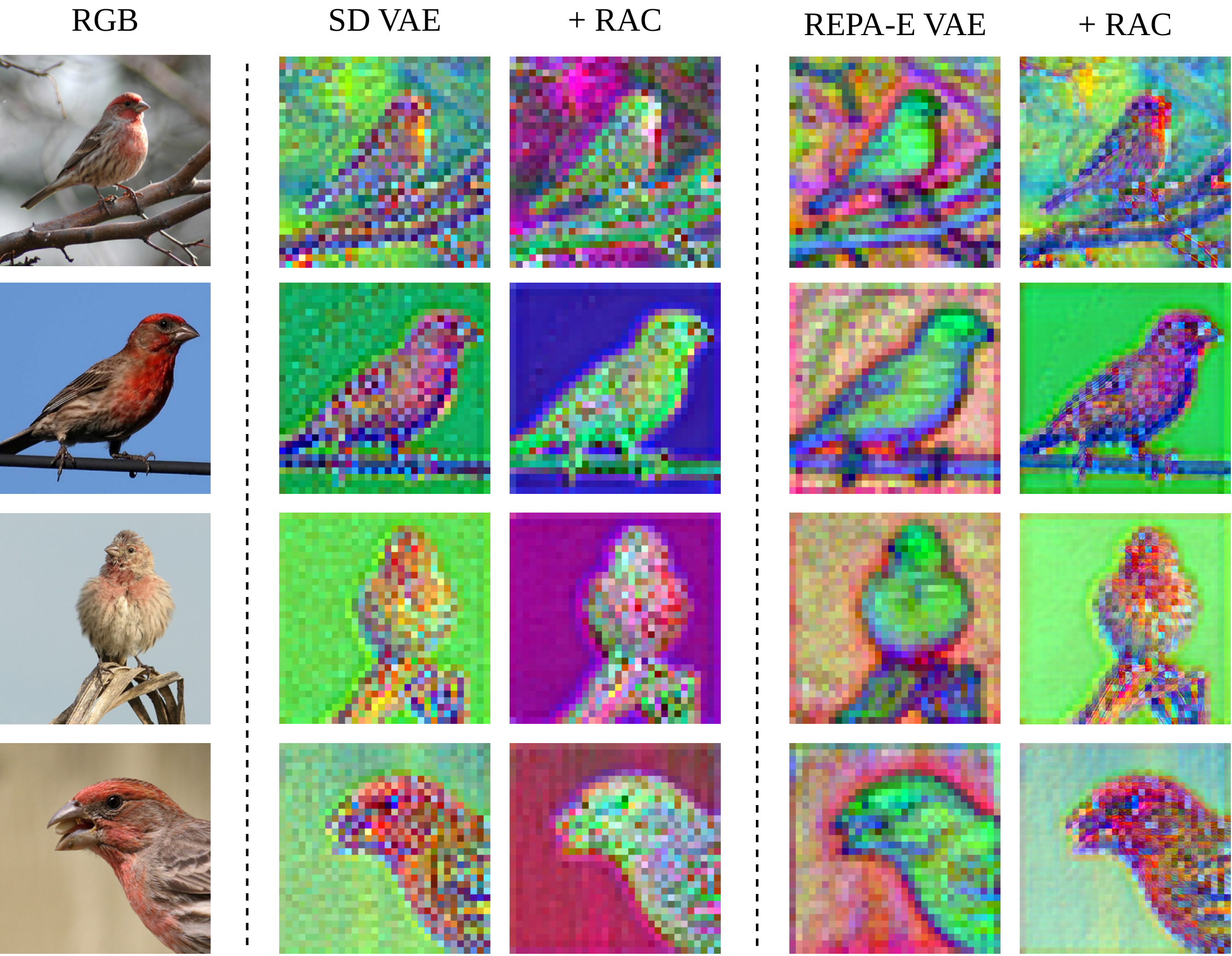}
        \vskip 0.05in
        \captionsetup{width=\linewidth}
        \caption{{PCA Analysis on VAE Latent Space Structure. RAC yields cleaner, more organized latent structure.}}
    \end{subfigure}%
    }
    \hfill
    \begin{subfigure}[t]{0.44\textwidth}
        \centering
        \raisebox{-4pt}{\includegraphics[width=0.98\textwidth]{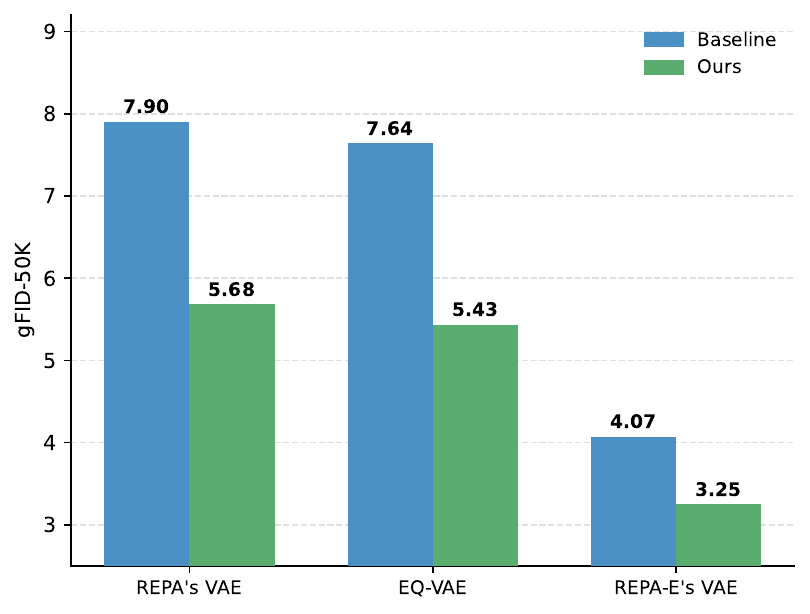}}
        \vspace{-4pt}
        \caption{{Performance Improvements with Baseline (400K Steps)}}
    \end{subfigure}
    \hfill
\vskip -0.05in
\caption{\textbf{{The State Space of RAC is a More Organized Representation Space.}}
\textcolor{purple}{(a)} We used PCA to conduct a visual analysis of the state space that we encoded. This state space represents the state before the final output of the encoder. We can clearly observe that the training of RAC has made the original Latent representation more orderly and clean. \textcolor{purple}{(b)} Comparison of gFID-50K on ImageNet 256×256 \cite{imgnet} across different VAE configurations. We report results for the baseline methods (REPA \cite{repa}, REPA-E \cite{leng2025repaeunlockingvaeendtoend,Rombach_2022_CVPR} and EQ-VAE \cite{eqvae}), comparing each against our approach trained under identical settings. Lower is better. RAC consistently outperforms the corresponding baselines across all VAE settings.
}
\vskip -0.2in
\label{fig:pca}
\end{figure*}

\paragraph{Reconstruction Quality Comparison.} As shown in Figure~\ref{fig:pca}(a), RAC-induced latent representations exhibit significantly cleaner, more coherent structures than the standard SD-VAE, which stem from the path consistency and latent alignment objectives jointly regularizing the state space during training. This structural regularity not only benefits generation—as reflected by the consistent gFID-50K improvements across all VAE configurations in Figure~\ref{fig:pca}(b)—but also yields more faithful image reconstruction, since a well-organized state space provides smoother and more stable ODE trajectories during the reverse decoding process. Quantitatively, RAC achieves competitive reconstruction quality while improving generative performance, demonstrating that the two objectives are complementary rather than conflicting within our flow-based autoencoding framework.

\begin{table}[t]
    \vspace{4pt}
    \centering
    \caption{\textbf{VAE architecture and decoder scale on ImageNet-1K.} Left: generation quality across VAE backbones and training configurations. Right: reconstruction quality (rFID) and computation (GFLOPs) across RAC decoder scales. RAC consistently improves quality while reducing parameters or computation. Default settings are highlighted in \colorbox{gray!20}{gray}. See App.~\ref{sec:latency_cost} for generation.
    }
    \label{tab:different_vae}
    \vspace{-12pt}
    \footnotesize
    \setlength{\tabcolsep}{2.2mm}{
    \resizebox{\linewidth}{!}{
    \begin{tabular}{cc}
    \begin{tabular}[t]{lcccccc}
        \toprule
        \textbf{Autoencoder} & \textbf{gFID}$\downarrow$ & \textbf{sFID}$\downarrow$ & \textbf{IS}$\uparrow$ & \textbf{Prec.}$\uparrow$ & \textbf{Rec.}$\uparrow$ & \textbf{$\Delta$Params} \\
        \midrule
        \rowcolor{white}SD-VAE~\cite{Rombach_2022_CVPR} & 24.1 & 6.25 & 55.7 & 0.62 & 0.60 & 0.0\% \\
        \rowcolor{white}+REPA-E \cite{leng2025repaeunlockingvaeendtoend} & 16.3 & 5.69 & 75.0 & 0.68 & 0.60 & 0.0\% \\
        \rowcolor{blue!8}\textbf{+RAC (Ours)} & \textbf{14.7} & \textbf{5.51} & \textbf{78.4} & \textbf{0.70} & \textbf{0.61} & \textbf{-40.8\%} \\
        \midrule
        \rowcolor{white}IN-VAE \cite{Rombach_2022_CVPR,leng2025repaeunlockingvaeendtoend} & 22.7 & 5.47 & 56.0 & 0.62 & 0.62 & 0.0\% \\
        \rowcolor{white}+REPA-E \cite{leng2025repaeunlockingvaeendtoend} & 12.7 & 5.57 & 84.0 & 0.69 & 0.62 & 0.0\% \\
        \rowcolor{blue!8}\textbf{+RAC (Ours)} & \textbf{11.2} & \textbf{5.21} & \textbf{87.5} & \textbf{0.71} & \textbf{0.63} & \textbf{-40.7\%} \\
        \midrule
        \rowcolor{white}VA-VAE~\cite{yao2025vavae} & 12.8 & 6.47 & 83.8 & 0.71 & 0.58 & 0.0\% \\
        \rowcolor{white}+REPA-E \cite{leng2025repaeunlockingvaeendtoend} & 11.1 & 5.31 & 88.8 & 0.72 & 0.61 & 0.0\% \\
        \rowcolor{blue!8}\textbf{+RAC (Ours)} & \textbf{9.8} & \textbf{5.08} & \textbf{91.4} & \textbf{0.73} & \textbf{0.62} & \textbf{-40.7\%} \\
        \bottomrule
    \end{tabular}
    &
    \hspace{6pt}
    \begin{tabular}[t]{lcc}
        \toprule
        Decoder & rFID & GFLOPs \\
        \midrule
        \rowcolor{white}SD-VAE-0.1x & 0.44 & 94.3 \\
        \rowcolor{white}SD-VAE-0.2x & 0.39 & 152.4 \\
        \rowcolor{white}SD-VAE-0.3x & 0.35 & 254.5 \\
        \midrule
        \rowcolor{gray!20}SD-VAE & \gc{0.62} & \gc{310.4} \\
        \bottomrule
        \multicolumn{3}{p{4.6cm}}{\cellcolor{white}\vspace{4pt}\textbf{Scalable Decoders:} improve rFID while remaining much more efficient than VAEs. 0.1x indicates that the parameter being used is one-tenth of the SD.}
    \end{tabular}
    \end{tabular}
    }}
    \vspace{-16pt}
\end{table}

Table~\ref{tab:different_vae} (left) presents a comprehensive comparison of generation quality across three VAE backbones—SD-VAE, IN-VAE, and VA-VAE—each evaluated under three training configurations: the vanilla baseline, +REPA-E, and our proposed +RAC. Across all metrics including gFID, sFID, IS, Precision, and Recall, RAC consistently achieves the best performance within each VAE group. For SD-VAE, RAC reduces gFID from 24.1 (baseline) to 14.7, outperforming the +REPA-E variant at 16.3, while also improving IS from 75.0 to 78.4; a similar trend holds for IN-VAE. For the strongest backbone VA-VAE, RAC further pushes gFID down to 9.8 from 11.1, with sFID improving from 5.31 to 5.08 and IS reaching 91.4. Finally, we presented the parameter changes of our RAC compared to the original VAE. It can be observed that the parameters have generally decreased by approximately 41\%.
These gains across all three VAE configurations demonstrate that RAC
provides a complementary improvement to any pretrained VAE backbone, reinforcing the effectiveness of our flow-based autoencoding framework as a plug-in enhancement for latent training.

Table~\ref{tab:different_vae} (right) studies the effects of decoder scale on reconstruction quality (rFID) and computational efficiency (GFLOPs) in RAC. We evaluate three scaled variants—SD-VAE-0.1x, SD-VAE-0.2x, and SD-VAE-0.3x—whose factors denote parameter fractions relative to the full SD-VAE decoder used as reference. Despite retaining only a fraction of its parameters, all three achieve substantially better rFID.
This counterintuitive but important result shows that, within RAC, a significantly smaller decoder both reduces computation (by approximately 70\%) and improves reconstruction fidelity. It suggests that continuous flow-based decoding alleviates the representational burden on decoder capacity, enabling high-quality reconstruction under aggressive parameter reduction.

\begin{figure*}[t]
    \begin{center}
    \centerline{\includegraphics[width=0.95\linewidth]{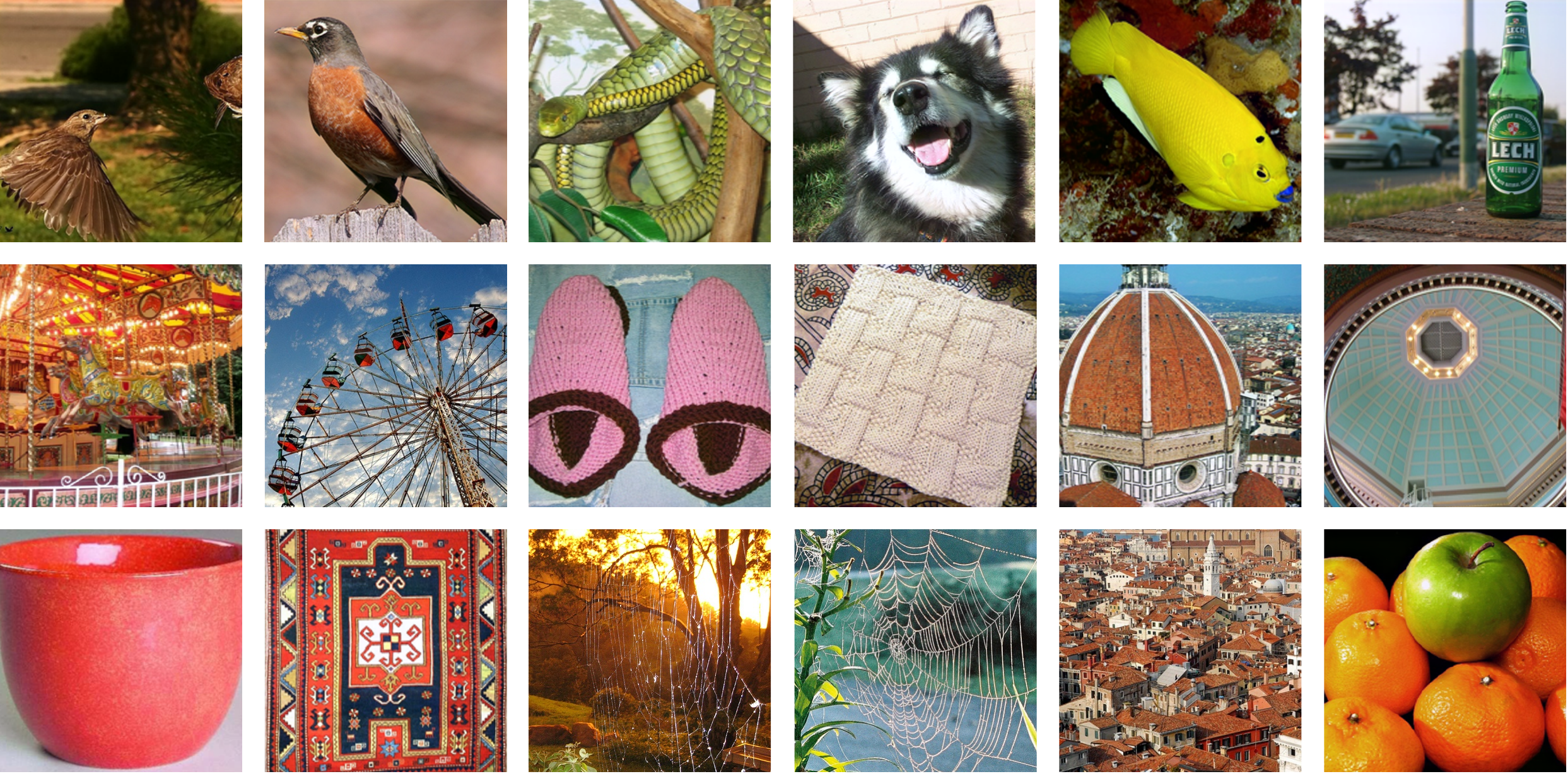}}
    \caption{\textbf{Qualitative Results on Imagenet 256 $\times$ 256} using RAC. It can achieve significant results in the state reconstruction mode with just 30k steps.}
    \label{fig:viz-results}
    \end{center}
    \vspace{-24pt}
\end{figure*}

\vspace{-12pt}
\paragraph{Visual Reconstruction Showcase.} Figure~\ref{fig:viz-results} shows high-fidelity RAC reconstructions across animals, textures, architecture, and everyday objects on ImageNet 256$\times$256 after only 30k training steps. RAC preserves fine-grained details—dog fur, snake scales, woven slippers, and ornate carpet patterns—that are challenging for conventional VAE decoders under aggressive spatial compression. It also maintains global structure and color, evident in the robin's sharp silhouette, the beer bottle's vivid green, and the Florence Cathedral's dome geometry.\looseness=-1

\begin{wrapfigure}{l}{0.38\textwidth}
  \centering
  \vspace{-16pt}
  \includegraphics[width=0.38\textwidth]{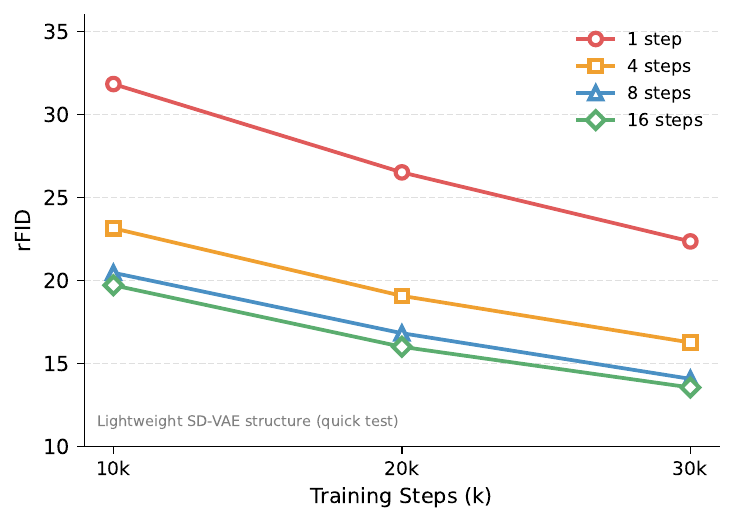}
  \vspace{-16pt}
  \caption{\textbf{rFID vs.\ Training Steps under Different Inference Steps.} Results on a lightweight SD-VAE structure for 1, 4, 8, and 16 decoding steps. Lower is better.}
  \label{fig:rfid_line}
  \vspace{-16pt}
\end{wrapfigure}
\vspace{-12pt}
\paragraph{Decoding Steps vs. Reconstruction Quality.} Figure~\ref{fig:rfid_line} investigates how the number of inference steps in the RAC decoding process affects reconstruction quality (rFID) over the course of training. We conduct this analysis on a lightweight SD-VAE variant as a controlled ablation, evaluating four inference step configurations: 1-16 steps, measured at 10-30k training steps. As expected, increasing the number of decoding steps consistently improves rFID across all training checkpoints, with the gap between 1-step and 16-step decoding being most pronounced at early training stages. By 30k steps, rFID improves from 22.36 at 1 step down to 13.56 at 16 steps, demonstrating that the flow-based decoder benefits substantially from finer ODE integration, with clear separation between curves.
The ordering remains consistent across checkpoints, confirming that additional decoding steps improve reconstruction quality.\looseness=-1

\begin{figure}[t]
    \begin{center}  \centerline{\includegraphics[width=\linewidth]{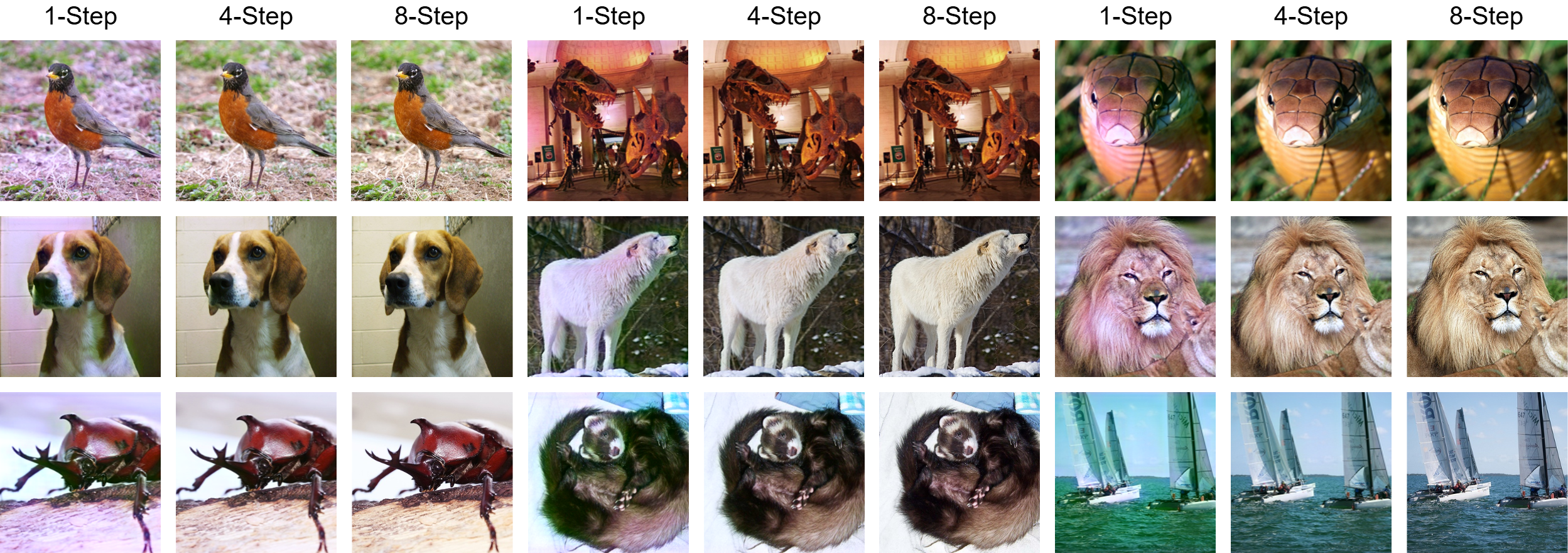}}
    \captionsetup{font=footnotesize,skip=4pt}
    \caption{\textbf{Generative Decoding under Ultra-short Training:} To visually demonstrate the benefits of our generative decoding on the results, we only trained for 1k steps (this is merely 1/10k of the full training duration in other works), and then we systematically observed the number of decoding inferences impacts performance. It's clearly see that during the initial training phase, multi-step decoding significantly contributed to the improvement in quality compared to single-step baselines.
    }
    \label{fig:rac_gen_qualitative}
    \end{center}
    \vspace{-24pt}
\end{figure}

\begin{figure}[t]
  \centering
\begingroup
\definecolor{racLatent}{HTML}{DDE7FA}
\definecolor{racImage}{HTML}{F9E2CB}
\definecolor{racSeparate}{HTML}{EBDFF0}
\definecolor{racShared}{HTML}{D5E7D9}
\definecolor{racDecode}{HTML}{AF4C2D}
\definecolor{racNote}{HTML}{584674}
\begin{tikzpicture}[
  x=1pt,y=1pt,
  font=\sffamily\fontsize{9}{10}\selectfont,
  every node/.style={inner sep=0pt,outer sep=0pt},
  racBox/.style={draw=black!75,line width=0.65pt,rounded corners=2pt},
  racArrow/.style={-{Latex[length=3.5pt,width=3.5pt]},line width=0.9pt},
  racLabel/.style={font=\sffamily\fontsize{9}{10}\selectfont}
]
  \path[use as bounding box] (0,0) rectangle (396,81);

  \node[racBox,fill=racLatent,minimum width=14pt,minimum height=14pt]
    (decLatent) at (20,59) {};
  \node[racBox,fill=racSeparate,minimum width=68pt,minimum height=22pt,
        font=\sffamily\bfseries\fontsize{9}{10}\selectfont]
    (decoder) at (92,59) {Decoder Net};
  \node[racBox,fill=racImage,minimum width=24pt,minimum height=24pt]
    (decImage) at (168,59) {};
  \draw[racArrow] (decLatent.east) -- (decoder.west);
  \draw[racArrow] (decoder.east) -- (decImage.west);
  \node[racLabel] at (20,42) {latent};
  \node[racLabel] at (168,40) {image};

  \node[racBox,fill=racImage,minimum width=24pt,minimum height=24pt]
    (encImage) at (20,20) {};
  \node[racBox,fill=racSeparate,minimum width=68pt,minimum height=22pt,
        font=\sffamily\bfseries\fontsize{9}{10}\selectfont]
    (encoder) at (92,20) {Encoder Net};
  \node[racBox,fill=racLatent,minimum width=14pt,minimum height=14pt]
    (encLatent) at (168,20) {};
  \draw[racArrow] (encImage.east) -- (encoder.west);
  \draw[racArrow] (encoder.east) -- (encLatent.west);
  \node[racLabel] at (20,1) {image};
  \node[racLabel] at (168,3) {latent};

  \node[racBox,fill=racLatent,minimum width=24pt,minimum height=24pt]
    (sharedLatent) at (217,51) {};
  \node[racBox,fill=racShared,minimum width=74pt,minimum height=30pt,
        align=center,font=\sffamily\bfseries\fontsize{9}{10}\selectfont]
    (shared) at (297,51) {RAC\\(bidirectional)};
  \node[racBox,fill=racImage,minimum width=24pt,minimum height=24pt]
    (sharedImage) at (375,51) {};
  \draw[racArrow,racDecode] (229,55) -- (260,55);
  \draw[racArrow,racDecode] (334,55) -- (363,55);
  \draw[racArrow] (260,47) -- (229,47);
  \draw[racArrow] (363,47) -- (334,47);
  \node[racLabel,text=racDecode] at (297,75) {decode reverse = False};
  \node[racLabel] at (217,32) {latent\_full};
  \node[racLabel] at (375,32) {image state};
  \node[racLabel] at (297,25) {encode reverse = True};
  \node[racLabel,text=racNote] at (297,12) {one backbone, shared weights theta};
  \node[racLabel,text=racNote] at (297,1) {effective params \textasciitilde\ P};

  \foreach \racX/\racY in {20/59,168/20} {
    \draw[black!18,line width=0.3pt]
      (\racX-4,\racY) -- (\racX+4,\racY)
      (\racX,\racY-4) -- (\racX,\racY+4);
  }
  \foreach \racX/\racY in {168/59,20/20,217/51,375/51} {
    \draw[black!14,line width=0.25pt,step=4pt]
      (\racX-8,\racY-8) grid (\racX+8,\racY+8);
  }
\end{tikzpicture}%
\endgroup
  \captionsetup{font=footnotesize,skip=4pt}
  \caption{\textbf{Parameter Efficiency of Bidirectional Design.} A conventional
    VAE-style bidirectional model uses separate encoder and decoder backbones to map
    between compact latent and image spaces. In contrast, RAC uses a shared
    bidirectional model in the full-resolution state space, achieving encoding and
    decoding by reversing the flow direction rather than duplicating the backbone.}
  \label{fig:VAE_vs_RAC}
  \vspace{-8pt}
\end{figure}
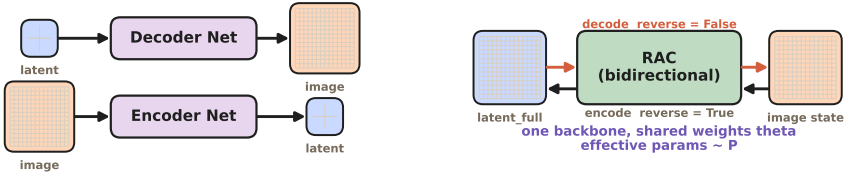

\subsection{Generation Quality}
\label{sec:gen}

\paragraph{Generated Image Samples.}
Figure~\ref{fig:rac_gen_qualitative} illustrates generative decoded image samples under ultra-short training (1k steps, 1/10k of full training duration), comparing 1-8 step generative decoding across diverse ImageNet categories. Even at this early training stage, the multi-step decoding of RAC progressively refines image quality: 1-step outputs capture coarse structure and semantics, while 4-step and 8-step inferences recover finer textures, sharper edges, and more faithful color rendition. This behavior demonstrates that RAC's generative decoder functions as an iterative refinement mechanism, effectively compensating for the limited encoder training by leveraging the diffusion prior to denoise and enhance latent representations at inference time.\looseness=-1

\begin{wraptable}{r}{0.52\textwidth}
    \vspace{-10pt}
    \centering
    \caption{\textbf{Model scale comparison on ImageNet 256$\times$256.} Our RAC improves generation quality across all scales, achieving consistent gFID reductions over both the vanilla baseline and REPA-E.
    }
    \label{tab:different_size}
    \resizebox{\linewidth}{!}{
    \footnotesize
    \setlength{\tabcolsep}{2.2mm}{
    \begin{tabular}{lccccc}
        \toprule
        \textbf{Diff. Model} & \textbf{gFID}$\downarrow$ & \textbf{sFID}$\downarrow$ & \textbf{IS}$\uparrow$ & \textbf{Prec.}$\uparrow$ & \textbf{Rec.}$\uparrow$ \\
        \midrule
        SiT-B (130M) & 49.5 & 7.00 & 27.5 & 0.46 & \textbf{0.59} \\
        +REPA-E & 34.8 & 6.31 & 39.1 & 0.57 & 0.59 \\
        \rowcolor{blue!8} \textbf{+RAC (Ours)} & \textbf{32.1} & \textbf{6.18} & \textbf{41.3} & \textbf{0.59} & \textbf{0.60} \\
        \midrule
        SiT-L (458M) & 24.1 & 6.25 & 55.7 & 0.62 & \textbf{0.60} \\
        +REPA-E & 16.3 & 5.69 & 75.0 & 0.68 & 0.60 \\
        \rowcolor{blue!8} \textbf{+RAC (Ours)} & \textbf{14.7} & \textbf{5.51} & \textbf{78.4} & \textbf{0.70} & \textbf{0.61} \\
        \midrule
        SiT-XL (675M) & 19.4 & 6.06 & 67.4 & 0.64 & \textbf{0.61} \\
        +REPA-E & 12.8 & 5.04 & 88.8 & 0.71 & 0.58 \\
        \rowcolor{blue!8} \textbf{+RAC (Ours)} & \textbf{11.2} & \textbf{4.87} & \textbf{92.5} & \textbf{0.73} & \textbf{0.61} \\
        \midrule
    \end{tabular}
    }}
    \vspace{-8pt}
\end{wraptable}
\paragraph{Generation Quality Comparison.}
Table~\ref{tab:different_size} presents a quantitative comparison of RAC against baseline SiT models \cite{sit} and REPA-E \cite{leng2025repaeunlockingvaeendtoend}.
RAC consistently achieves lower gFID and sFID scores while improving Inception Score (IS) and Precision across all scales, demonstrating that the rectified autoencoder with consistency path provides complementary benefits across representation-aligned generative frameworks. Specifically, RAC reduces gFID by 2.7, 1.6, and 1.6 points over REPA-E for SiT-B, SiT-L, and SiT-XL respectively, while maintaining competitive Recall. These results confirm that RAC's improvements are not model-capacity-specific but scale consistently, making it a broadly applicable enhancement to rectified flow-based generative frameworks.


\subsection{Bidirectional Consistency}
\label{sec:bidir}

\begin{figure}[t]
  \centering
  \includegraphics[width=0.90\linewidth]{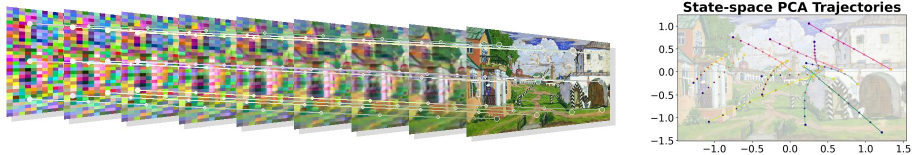}
  \captionsetup{font=footnotesize,skip=4pt}
  \caption{\textbf{Encoding and Decoding Trajectory Symmetry.}
  Left: RAC decodes the full-resolution state to the image through a sequence of intermediate states. Right: the trajectories of sampled spatial positions are projected to a 2D PCA space, revealing that the decoding process follows structured and coordinated paths in the learned state space rather than independent pixel-wise changes.}
  \label{fig:trajectory_perspective-pca}
  \vspace{-8pt}
\end{figure}

\paragraph{Parameter Efficiency of Bidirectional Design.}
As illustrated in Fig.~\ref{fig:VAE_vs_RAC}, a conventional VAE-style bidirectional design relies on two separate networks: a decoder that maps a compact latent to image space and an encoder that maps the image back to the compact latent. This design achieves bidirectionality by duplicating the backbone across two directions. In contrast, RAC uses a single shared bidirectional model operating in the full-resolution state space. Encoding and decoding are both realized by the same model, with the transformation direction controlled by reversing the flow direction. This shared design avoids the need for a dedicated encoder-decoder pair and therefore provides a substantially more parameter-efficient bidirectional formulation.\looseness=-1

\vspace{-12pt}
\paragraph{Encoding and Decoding Trajectory Symmetry.} As shown in Fig.~\ref{fig:trajectory_perspective-pca}, RAC does not transform the representation by a single abrupt jump from latent-like initialization to the final image. Instead, the model evolves the sample through a sequence of full-resolution intermediate states, producing a smooth decoding trajectory in the shared state space. The left panel visualizes this evolution directly in image space, where the representation gradually acquires semantic structure and local detail over time. The right panel further projects the trajectories of sampled spatial positions into a 2D PCA space. Although each sampled position corresponds to a fixed spatial location, its feature vector follows a coherent trajectory rather than remaining static or changing independently.
Our model does not require two unrelated transformations for encoding and decoding, but instead traverses a common representation manifold in opposite directions.\looseness=-1

\begin{figure}[t]
\begin{minipage}{0.64\columnwidth}
  \centering
  \vspace{0pt}
  \includegraphics[width=0.92\linewidth]{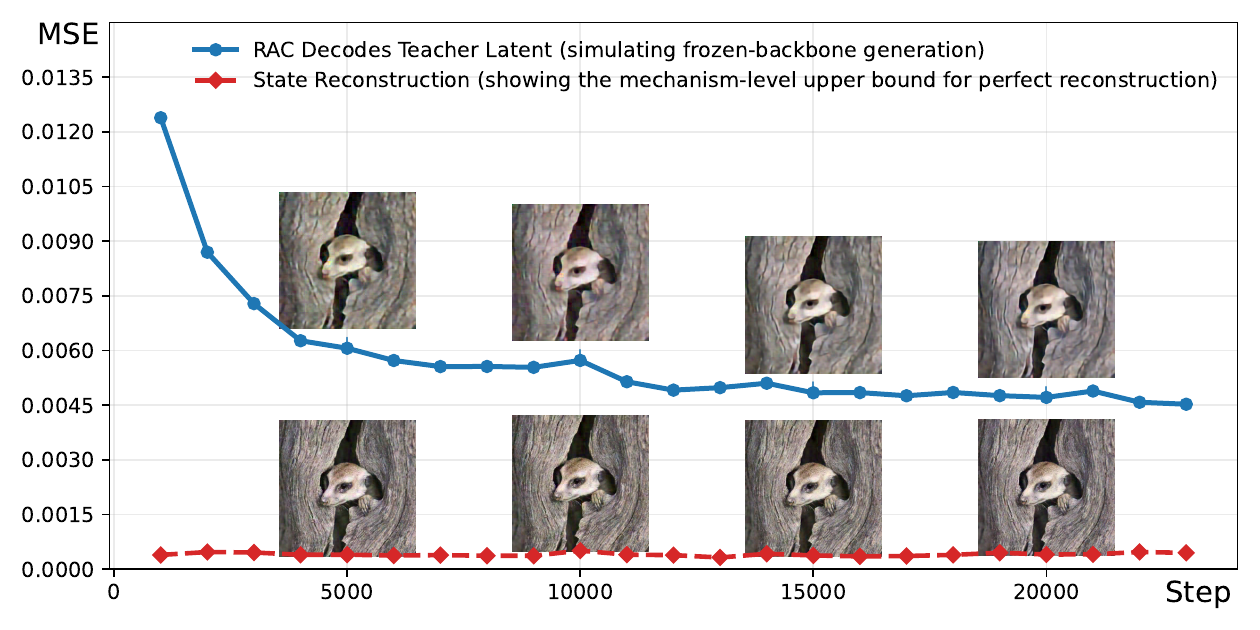}
\end{minipage}\hspace*{0.02\columnwidth}%
\begin{minipage}{0.34\columnwidth}
  \centering
  \vspace{5pt}
  \scriptsize
  \includegraphics[width=0.92\linewidth]{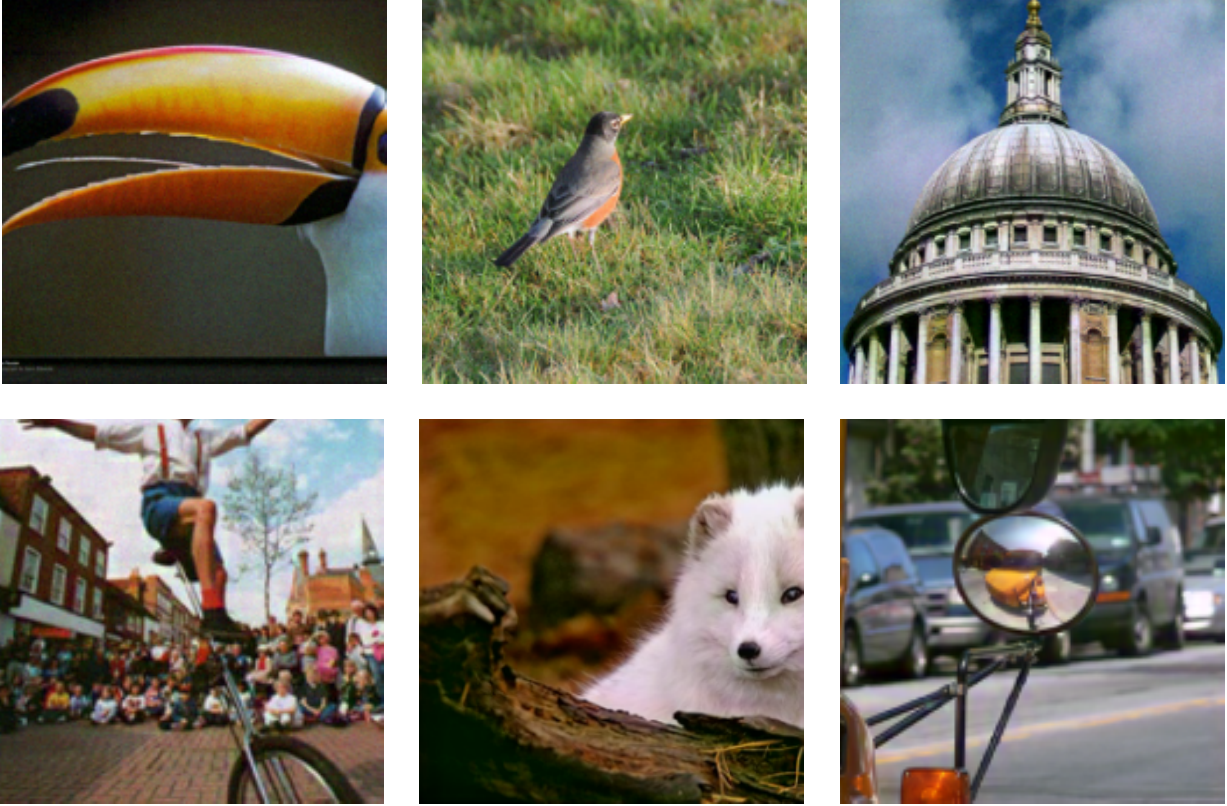}\\[-2pt]
  \colorbox{white}{%
  \parbox{\dimexpr\linewidth-2\fboxsep\relax}{\raggedright \scriptsize \textbf{Our learning mechanism is unique:} %
  With only 100 training steps, the state reconstruction achieves extremely high quality.}%
}
\end{minipage}
\captionsetup{font=footnotesize,skip=4pt}
\caption{\textbf{Mechanism-level ability of RAC.} Left: reconstruction errors during training. Decoding teacher latents with RAC simulates frozen-backbone generation, while state reconstruction measures the upper bound of reconstruction within RAC's shared full-resolution state space. Right: qualitative state reconstruction examples.\looseness=-1}
\label{fig:perfect_re}
\vspace{-8pt}
\end{figure}

\subsection{Generation--Reconstruction Gap Analysis}
\label{sec:gap}

\begin{wraptable}{R}{0.42\textwidth}
    \vspace{-8pt}
    \centering
    \captionsetup{font=footnotesize,skip=4pt}
    \caption{\textbf{Training and decoding-step ablation:} We compare training configurations and decoding steps under 100K iterations of end-to-end training from scratch, evaluating their contributions to generation quality.}
    \label{tab:abl}
    \resizebox{0.41\textwidth}{!}{
    \footnotesize
    \setlength{\tabcolsep}{2mm}{
    \begin{tabular}{lccccc}
        \toprule
        \textbf{Method} & \textbf{gFID}$\downarrow$ & \textbf{sFID}$\downarrow$ & \textbf{IS}$\uparrow$ & \textbf{Prec.}$\uparrow$ & \textbf{Rec.}$\uparrow$ \\
        \midrule
        \multicolumn{6}{c}{\textbf{100K Iterations (20 Epochs)}} \\
        \midrule
        REPA~\cite{repa} & 19.40 &  6.06 & 67.4 & 0.64 & \textbf{0.61} \\
        \rowcolor{yellow!10} \textbf{REPA-E} (scratch) & 14.12 & 7.87 & 83.5 & 0.70 & 0.59 \\
        \rowcolor{blue!8} \textbf{RAC} (scratch, 1-step) & 14.05 & 7.83 & 83.9 & 0.70 & 0.61 \\
        \rowcolor{blue!15} \textbf{RAC} (scratch, 4-step) & \textbf{13.47} & \textbf{7.31} & \textbf{86.2} & \textbf{0.71} & 0.61 \\
        \bottomrule
    \end{tabular}
    }}
    \vspace{-16pt}
\end{wraptable}

\paragraph{Generation--Reconstruction Gap Analysis.} As shown in Figure~\ref{fig:perfect_re}, RAE \cite{zheng2026diffusion} and Flux vae \cite{flux, labs2025flux1kontextflowmatching} achieve better results by increasing the number of channels and enhancing the expression ability of variables. However, if the state representation is used for training, it can learn a high-dimensional space without information loss. At this point, it is almost equivalent to fixed reconstruction learning generation. We can clearly see in the line graph that the generation ability is weaker than the reconstruction. This directly confirms our assumption in the Introduction and indicates that we have discovered in the mechanism how to unify generation and reconstruction with a training objective.\looseness=-1

\vspace{-12pt}
\paragraph{Quantitative Gap between Generation and Reconstruction.} As shown in Figure~\ref{fig:perfect_re} (Right), with just 100 steps, our state reconstruction is nearly perfect. In practice, using encoded state condensed variables equates to a traditional variable-size compression paradigm. This robust behavior occurs because our training methodology constructs a high-dimensional continuous space mapped directly from the raw image, which subsequently stabilizes near the original pixel dimension. Consequently, even though this latent space is imperfect, it does not degrade reconstruction performance.\looseness=-1

\subsection{Ablation Study \& Latent Space Analysis}
\label{sec:ablation}
\label{sec:visual}

\paragraph{Ablation Study on RAC Components.} Table~\ref{tab:abl} suggests that even minimal RAC consistency regularization preserves latent quality. Increasing the integration from 1 to 4 steps further improves gFID, sFID, IS, and Precision, showing that the trajectory contributes measurable refinement beyond the shared parameterization alone. The gains arise under the same training budget and initialization, indicating that additional flow steps expose a more effective correction path for the learned latent space.\looseness=-1

\begin{figure}[t]
  \centering
  \captionsetup{font=footnotesize,skip=3pt}
  \captionsetup[subfigure]{font=footnotesize,skip=2pt}
  \begin{subfigure}[t]{0.377\textwidth}
    \centering
    \includegraphics[width=0.90\textwidth]{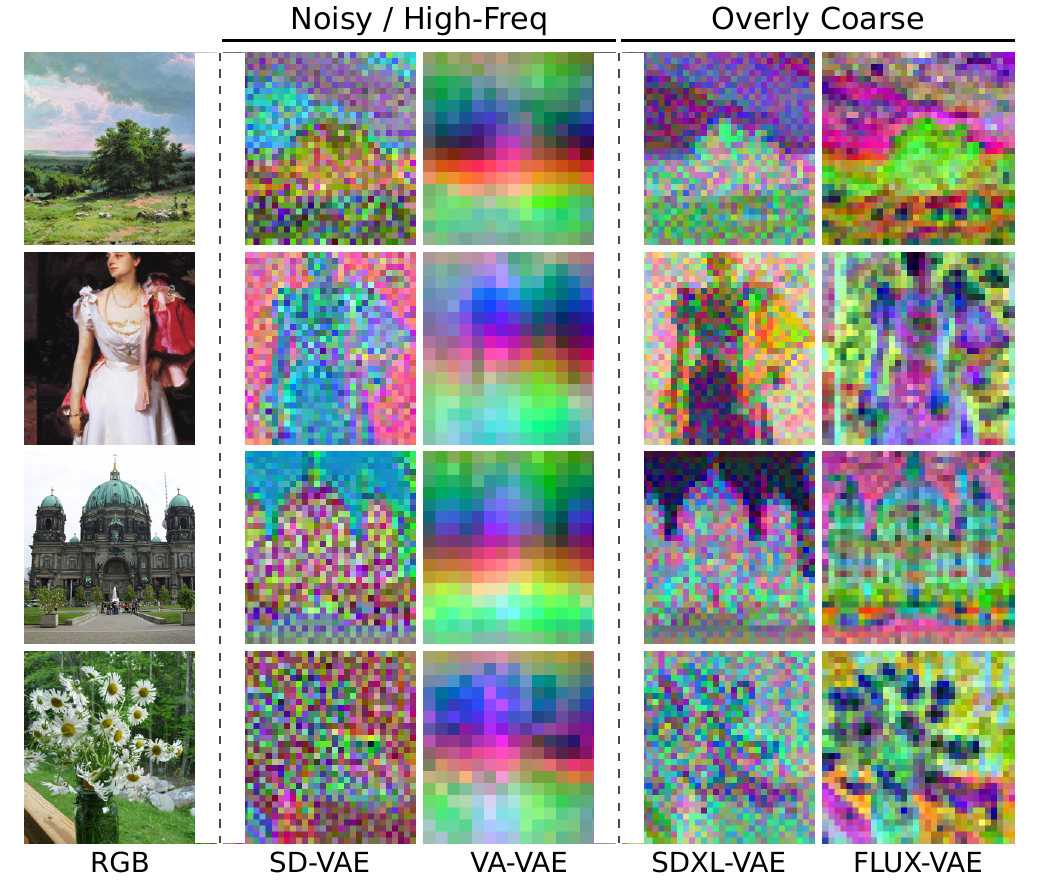}
    \caption{Latent artifacts in pretrained VAEs \citep{Rombach_2022_CVPR,sdxl,flux,yao2025vavae}}
    \label{fig:pretrained-vae-pca}
  \end{subfigure}
  \hfill
  \begin{subfigure}[t]{0.523\textwidth}
    \centering
    \includegraphics[width=0.90\textwidth]{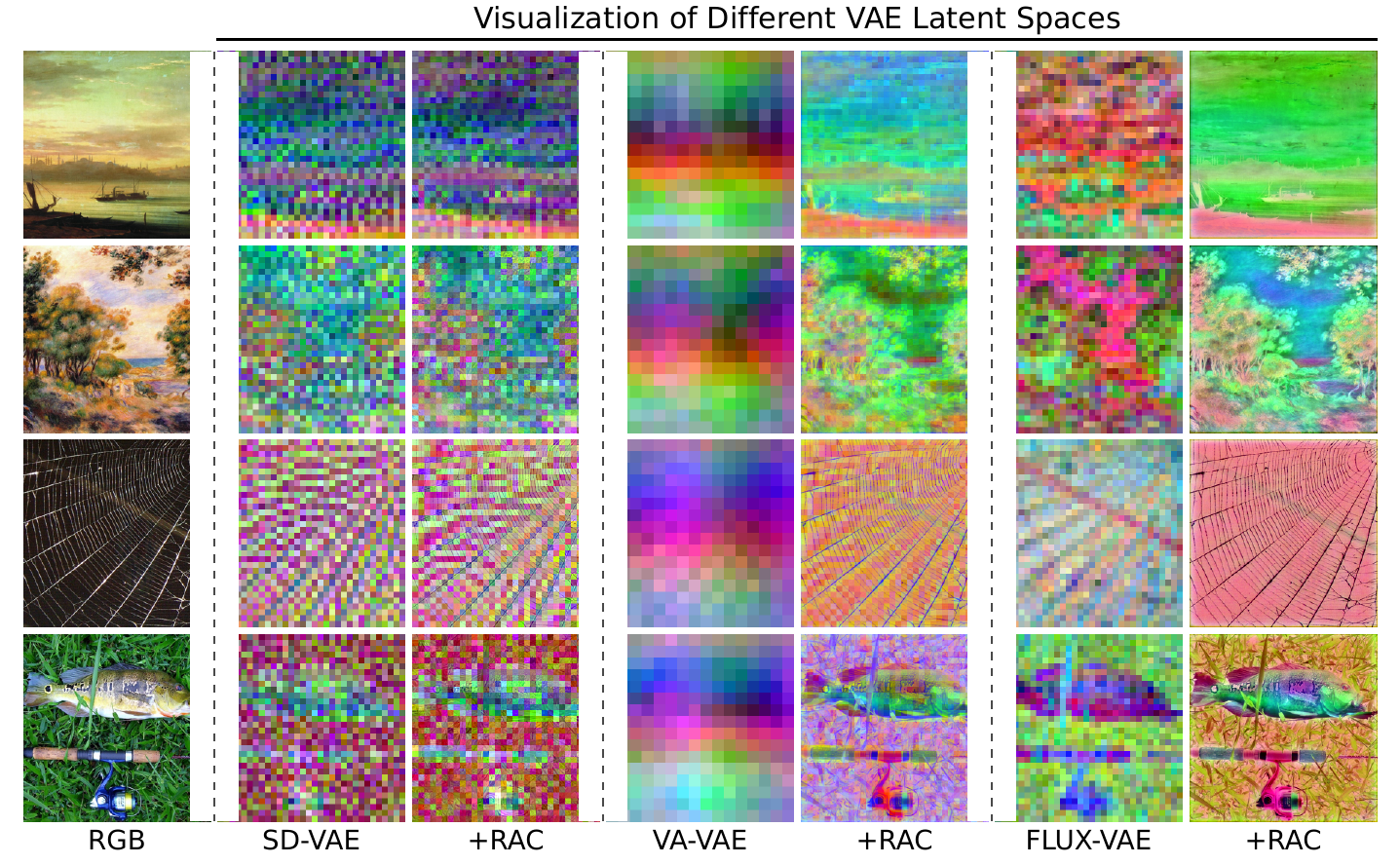}
    \caption{RAC yields cleaner backgrounds, richer structural details, and more stable colors, indicating improved spatial coherence in the learned latent representations.}
    \label{fig:repae-latentspace-pca}
  \end{subfigure}
  \caption{Baseline latent manifolds show architecture-dependent anisotropy, with principal components dominated by high-frequency texture or coarse low-frequency blocks. RAC reallocates variance to spatially coherent components, yielding cleaner, more content-aligned latent geometry while preserving scene identity.}
  \label{fig:latentspace-pca}
  \vspace{-12pt}
\end{figure}



\vspace{-12pt}
\paragraph{Latent Space PCA Visualization.} Figure~\ref{fig:latentspace-pca} highlights two complementary observations. First, different pretrained VAE architectures produce latent spaces with distinct but systematic defects: some are dominated by noisy high-frequency patterns, whereas others collapse fine details into coarse, low-information structures. Second, RAC-enhanced representations substantially regularize these latent distributions. Their spatially coherent principal components align with the consistent gFID and sFID improvements reported across the VAE backbones, linking the visual structure of the latent space to measurable generation benefits. This cross-architecture regularity shows that RAC provides a reliable refinement mechanism for pretrained representations.\looseness=-1


\vspace{-4pt}
\section{Conclusion}
This paper introduces RAC (Rectified Flow Auto Coder), an automatic encoding framework based on rectified flow, aiming to fundamentally solve the problem of inconsistency between generation and reconstruction in traditional VAEs. Unlike traditional VAEs which treat decoding as a single-step mapping, RAC models the decoding process as a continuous-time velocity field integration, enabling the model to gradually correct latent variables along the decoding path, thereby significantly reducing the performance gap between generation and reconstruction. Thanks to the time reversal mechanism, RAC naturally supports bidirectional inference, that is, the same velocity field model can simultaneously perform encoding and decoding functions, reducing the parameter quantity by nearly 41\%. 
Experimental results show that RAC outperforms existing SOTA VAEs in terms of both reconstruction and generation performance, even under aggressive 10× parameter reduction.\looseness=-1


\subsection*{AI use statement}
In this work, generative AI tools were used to polish the manuscript prose and assist with preparing explanatory illustrations. Generative AI tools were not used to design the model. All AI-assisted text and visual materials were reviewed and edited by the authors. The authors take responsibility for the final content of this paper, including all text, claims, and artifacts produced with AI assistance.

\bibliography{ref/main,ref/sf,ref/rae,ref/repa}
\bibliographystyle{iclr2027_conference}

\clearpage
\appendix

\section{Related Work}

\subsection{Flow-Based Generative Models}

Normalizing flow models \cite{zhai2025normalizing} learn invertible mappings between complex data distributions and simple priors.
Early works such as Real NVP~\cite{dinh2017density} and Glow~\cite{kingma2018glow} establish the foundation of flow-based generation by enabling tractable likelihood estimation through invertible transformations.
Recent studies extend flow models toward diffusion-inspired formulations. Flow Matching (FM)~\cite{lipman2023flow} and Rectified Flow (RF)~\cite{liu2023flow} reformulate diffusion training as direct vector field regression, enabling deterministic sampling and efficient training.
These methods bridge continuous normalizing flows~\cite{chen2018neural,fang2026streamflowtheoryalgorithmimplementation,song2023consistency} and diffusion models \cite{Rombach_2022_CVPR,salimans2022progressive}, achieving competitive generation quality with reduced sampling steps.

\subsection{Variational Autoencoders and Representation Learning}

Variational Autoencoders (VAEs)~\cite{kingma2014auto} provide a principled framework for learning latent-variable generative models via variational inference.
However, standard VAEs often suffer from posterior collapse and limited reconstruction fidelity. To alleviate these issues, Representations Autoencoders (RAEs)~\cite{zheng2026diffusion} replace KL regularization with explicit latent constraints.
Sparse Autoencoders (SAEs)~\cite{shu-etal-2025-survey} enforce sparsity priors to promote compact and interpretable representations.

Recent advances focus on improving latent expressiveness and generative performance.
REPA-E~\cite{leng2025repaeunlockingvaeendtoend} introduces residual posterior alignment for end-to-end latent diffusion training.
Rethinking VQ Tokenizers~\cite{Gu_2024_CVPR} identifies a competing relationship between reconstruction detail and semantic compression, and improves decoding while preserving the learned semantic latent space.
VQGAN~\cite{Esser_2021_CVPR} adopts vector-quantized autoencoders for efficient latent generative modeling.
Hybrid diffusion-autoencoder frameworks~\cite{yan2024hybridsdedgecloudcollaborative} further improve inference and generation consistency.
NVAE~\cite{NEURIPS2020_e3b21256} proposes hierarchical latent variables for scalable high-resolution synthesis. These approaches demonstrate that carefully designed latent regularization and architectural constraints are crucial for balancing reconstruction quality and generative diversity.

\subsection{Unified Modeling of Generation and Reconstruction}

Recent research explores unifying generation and reconstruction within a single probabilistic framework.
Latent diffusion models~\cite{Rombach_2022_CVPR} combine autoencoding architectures with diffusion processes for efficient high-resolution synthesis.

Several works interpret autoencoding and sampling as two sides of the same stochastic process.
Score-based models~\cite{song2021scorebased} and flow matching methods~\cite{lipman2023flow} naturally support both data reconstruction and generation via conditional sampling.
Consistency models~\cite{song2023consistency} further enable fast bidirectional mappings.

Recent end-to-end latent learning frameworks~\cite{leng2025repaeunlockingvaeendtoend,wang2026vaerepavariationalautoencoderrepresentation,Fang_2025_ICCV,11099440,fang2026stablesignerhierarchicalsign,fang2026signxcontinuoussignrecognition} explicitly couple inference and generation modules, improving representation alignment and robustness.
Such unified designs facilitate efficient downstream adaptation and controllable synthesis. Our work follows this line of research by jointly modeling reconstruction and generation within a unified flow-based latent framework, achieving improved fidelity and sampling efficiency.

\subsection{Generative Decoders and Diffusion Tokenizers}

A closely related line of work replaces deterministic autoencoder decoding with an iterative generative process. Epsilon-VAE~\cite{pmlr-v267-zhao25w} treats visual decoding as conditional denoising, while DiTo~\cite{chen2025diffusion} scales diffusion-based image tokenizers under a unified diffusion objective. SSDD~\cite{vallaeys2025ssdd} develops a diffusion decoder and distills it into an efficient single-step model. Earlier Diffusion Autoencoders~\cite{Preechakul_2022_CVPR} combine a semantic encoder with diffusion-based reconstruction, whereas FlexTok~\cite{pmlr-v267-bachmann25a} and SEMANTICIST~\cite{Wen_2025_ICCV} use generative decoders for structured one-dimensional tokenization. These works establish generative decoding as an effective alternative to a deterministic decoder, but retain an independent encoder for all or part of the representation.

\begin{table*}[!t]
  \centering
  \caption{\textbf{Mechanism-level comparison with generative-decoder and diffusion-tokenizer methods.} \emph{Full sharing} means that the complete encoding and decoding transformations use one parameter set; \emph{full reversal} means that the complete encoded state is obtained by reversing the same field used for decoding.}
  \label{tab:mechanism_comparison}
  \footnotesize
  \setlength{\tabcolsep}{2pt}
  \renewcommand{\arraystretch}{0.95}
  \resizebox{\textwidth}{!}{%
  \begin{tabular}{ccccccc}
    \toprule
    Method & \makecell[c]{Generative\\decoder} & \makecell[c]{Independent\\encoder} & \makecell[c]{Full\\sharing} & \makecell[c]{Full\\reversal} & \makecell[c]{Explicit round-trip\\loss} & \makecell[c]{Representation /\\compatibility} \\
    \midrule
    Epsilon-VAE~\cite{pmlr-v267-zhao25w} & \makecell[c]{Conditional\\diffusion} & Yes & No & No & No & \makecell[c]{Encoder-conditioned\\continuous latent} \\
    DiTo~\cite{chen2025diffusion} & \makecell[c]{Diffusion\\decoder} & Yes & No & No & No & \makecell[c]{Jointly learned\\continuous tokenizer} \\
    SSDD~\cite{vallaeys2025ssdd} & \makecell[c]{Diffusion +\\distillation} & Yes & No & No & No & KL-VAE replacement \\
    Diffusion Autoencoders~\cite{Preechakul_2022_CVPR} & DPM/DDIM & Semantic encoder & Partial & \makecell[c]{Stochastic\\code only} & No & \makecell[c]{Two-part semantic/\\stochastic code} \\
    FlexTok~\cite{pmlr-v267-bachmann25a} & Rectified flow & Yes & No & No & No & \makecell[c]{Variable-length\\1D tokens} \\
    SEMANTICIST~\cite{Wen_2025_ICCV} & \makecell[c]{Conditional\\diffusion} & Yes & No & No & No & \makecell[c]{Structured 1D\\causal tokens} \\
    \rowcolor{gray!15}\textbf{RAC (ours)} & \makecell[c]{\textbf{Rectified-flow}\\\textbf{VAE decoder}} & \makecell[c]{\textbf{Teacher only}\\\textbf{during training}} & \textbf{Yes} & \textbf{Yes} & \textbf{Yes} & \makecell[c]{\textbf{Pretrained KL-VAE}\\\textbf{latent spaces}} \\
    \bottomrule
  \end{tabular}}
\end{table*}

RAC converts a lightweight VAE decoder into a bidirectional velocity field whose complete encoding and decoding paths share parameters, and explicitly regularizes their reversal through round-trip consistency. The frozen teacher encoder supplies latent supervision during training. At inference, reverse-time integration uses the same RAC network for encoding. This design preserves compatibility with existing KL-VAE latent spaces while keeping the teacher encoder fixed.


\section{Implementation Details}
\label{sec:details}

\subsection{Algorithm and Implementation Details}
\label{sec:method_algorithm}
We use a KL-VAE teacher for latent supervision, AdamW for optimization, and mixed precision to reduce memory. The same parameters are shared for forward and reverse flows, which keeps the model compact while enforcing bidirectional structure. A random time grid and step noise are used to avoid overfitting to fixed integration schedules.

\paragraph{Architectural details.}
The learned latent projector is a four-layer convolutional network with a base channel width of 40 (approximately 100K parameters) that downsamples the encoded full-resolution state to the teacher latent size. The optional relative positional encoding is computed on normalized spatial coordinates $x,y\in[-1,1]$ as $\mathrm{pos}(x,y)=0.01[\cos(\pi x),\cos(\pi y)]$ and concatenated with the downsampled state and time channel.

\addtolength{\intextsep}{-12pt}
\begin{wrapalgorithm}{l}{0.5\textwidth}
  \footnotesize
  \caption{RAC Training (one iteration)}
  \label{alg:rac}
  \begin{algorithmic}[1]
    \Require image batch $\mathbf{x}$, teacher AE, steps $K$
    \State $\tilde{\mathbf{x}} \leftarrow 2\mathbf{x}-1$
    \State $\mathbf{z}_T \leftarrow \mathrm{Enc}_T(\tilde{\mathbf{x}})$
    \State $\mathbf{s}_0 \leftarrow \mathrm{expand}(\mathrm{pad}(\mathbf{z}_T))$
    \State $\{\mathbf{s}_k\}_{k=0}^{K} \leftarrow \mathrm{Flow}_\theta(\mathbf{s}_0)$
    \State $\mathbf{s}^\ast \leftarrow \mathrm{pad}(\tilde{\mathbf{x}})$
    \State compute $\mathcal{L}_{\text{recon}}, \mathcal{L}_{\text{path}}^{\text{dec}}$
    \State $\hat{\mathbf{s}}_0 \leftarrow \mathrm{Flow}_\theta^{-1}(\mathbf{s}^\ast)$
    \State $\hat{\mathbf{z}} \leftarrow \mathrm{down}(\hat{\mathbf{s}}_0)$
    \State compute $\mathcal{L}_{\text{latent}}, \mathcal{L}_{\text{pixel}}, \mathcal{L}_{\text{rt}}$
    \State $\mathcal{L} \leftarrow$ Eq.~\eqref{eq:rac_objective}; update $\theta$ with AdamW
  \end{algorithmic}
  \vspace{-4pt}
\end{wrapalgorithm}

\paragraph{Default settings.}
Unless otherwise noted, as shown in Algorithm~\ref{alg:rac}, we use $K{=}4$ Euler steps, $C_s{=}4$ state channels, AdamW with learning rate $3\times10^{-4}$ and $(\beta_1,\beta_2)=(0.9,0.99)$, input resolution $256$, and a small step noise ($\sigma=0.05$). We enable random time grids during training and use mixed precision when available. The weights are typically set to $\lambda_{\text{path}}=0.1$, $\lambda_{\text{latent}}=1.0$, $\lambda_{\text{pixel}}=1.0$, and $\lambda_{\text{rt}}=1.0$.

\paragraph{Teacher and latent protocol.}
The pretrained KL-VAE teacher encoder remains frozen throughout RAC training. During training, $\mathbf{z}_T=\mathrm{Enc}_T(\tilde{\mathbf{x}})$ provides a data-conditioned latent target. During generation, the attached backbone (e.g., SiT, DiT, or UNet) supplies $\mathbf{z}_T$ and determines its distribution. For a $256\times256$ image with an SD-VAE teacher, the latent size is typically $4\times32\times32$. Reported RAC training-step counts measure optimization of the velocity field and, when used, the lightweight latent projector; the teacher encoder remains fixed.

\paragraph{Operator and reversal details.}
The $\mathrm{pad}$ operator appends channels filled with $0.5$, placing the image and latent-derived endpoints in the same $C_s$-channel state space; with the default $C_s{=}4$, an RGB image receives one padding channel. The $\mathrm{expand}$ operator repeats each compact latent position over its corresponding spatial block, preserving all latent values in the full-resolution state. The $\mathrm{down}$ operator maps a reverse-integrated state back to the teacher latent size through the projector described above, and $\Pi$ selects the first three state channels as RGB. At every Euler step, the velocity field has the same $C_s\times H\times W$ shape as the evolving state. Encoding uses exactly the same parameters as decoding and reverses only the direction of Euler integration.

\paragraph{Why the teacher encoder is frozen.}
Joint optimization of an unconstrained encoder and RAC from scratch may learn an approximately invertible RGB shortcut that yields a weak latent for downstream generation. The high-dimensional intermediate trajectory can still carry information when the latent size is reduced. Freezing the pretrained encoder preserves an established latent space and isolates improvements due to RAC. This choice prioritizes representation stability and compatibility. Multi-step reverse integration requires more computation than one teacher-encoder pass.

\paragraph{Scaled variants and parameter accounting.}
The 0.1$\times$, 0.2$\times$, and 0.3$\times$ decoder variants uniformly adjust the channel widths of all layers to reach 10\%, 20\%, and 30\% of the full decoder parameter count, respectively, while preserving depth, upsampling stages, kernel sizes, and overall topology. This construction isolates decoder capacity, and each scale factor denotes the resulting total-parameter ratio relative to the full decoder. The reported 40.82\% reduction compares the parameter count of a conventional teacher encoder--decoder pair with RAC's single shared bidirectional model. The reduction concerns model parameters and deployment storage. Encoding latency is reported separately because reverse integration uses multiple steps.
\addtolength{\intextsep}{12pt}

\section{Additional Experiments}
\label{sec:additional_experiments}

\subsection{Comparison of Released Generative Decoders}
\label{sec:released_decoder_comparison}

We compare RAC with generative-decoder and tokenizer systems for which released pretrained
checkpoints can be evaluated through their native reconstruction pipelines. All methods receive the
same held-out inputs and use identical image preprocessing and metric implementations. Because each
released system retains its native latent representation, architecture, and inference schedule,
Table~\ref{tab:released_decoder_comparison} compares complete released systems under their native architectures, latent representations, and inference schedules.

\begin{table}[H]
  \centering
  \caption{\textbf{Pixel-fidelity comparison using released checkpoints.} All methods use the same held-out inputs and evaluation code. Decoder and total counts distinguish decoder parameters from the complete system; RAC shares one network across both directions.}
  \label{tab:released_decoder_comparison}
  \scriptsize
  \setlength{\tabcolsep}{4.5pt}
  \resizebox{\columnwidth}{!}{%
  \begin{tabular}{lccccc}
    \toprule
    Method & Decoder params. & Total params. & Decode NFE & PSNR $\uparrow$ & SSIM $\uparrow$ \\
    \midrule
    FlexTok-d12-d12~\cite{pmlr-v267-bachmann25a} & 172.45M & 341.65M & 20 & 17.152 & 0.4598 \\
    SEMANTICIST-L~\cite{Wen_2025_ICCV} & 457.05M & 609.76M & 250 & 22.199 & 0.6521 \\
    SSDD-S-F8C4~\cite{vallaeys2025ssdd} & 14.06M & 48.22M & 8 & 19.863 & 0.5126 \\
    \rowcolor{gray!15}\textbf{RAC-Tiny (ours)} & \textbf{1.46M} & \textbf{1.46M} & \textbf{4} & \textbf{23.917} & \textbf{0.7063} \\
    \bottomrule
  \end{tabular}}
\end{table}

Despite using only 1.46M shared parameters, RAC-Tiny obtains the strongest PSNR and SSIM among
the evaluated generative-decoder systems, improving PSNR over SSDD-S, FlexTok, and SEMANTICIST by
4.054, 6.765, and 1.718 dB, respectively. This performance is achieved with 33--417$\times$ fewer
total parameters than the compared generative-decoder systems.

\subsection{Robustness to Perturbed Latents}
\label{sec:latent_perturbation}

To directly test whether multi-step decoding is robust to inaccurate generation latents, we compare RAC with the single-step TAESD baseline under controlled latent perturbations. TAESD uses the latent produced by its frozen encoder, whereas RAC obtains its latent by reversing the shared velocity field for four steps and applying the learned compact projector. Both representations have size $4\times32\times32$. For each method, we perturb its native clean latent as
\begin{equation}
  \mathbf{z}_{\sigma}=\mathbf{z}_{\mathrm{native}}+\sigma\,\mathrm{std}(\mathbf{z}_{\mathrm{native}})\boldsymbol{\epsilon},
  \qquad \boldsymbol{\epsilon}\sim\mathcal{N}(\mathbf{0},\mathbf{I}).
\end{equation}
We evaluate 1,000 ImageNet images at $256\times256$ resolution, using the same four decoding steps as in RAC training. Differences are computed pairwise for each image before reporting their mean and 95\% confidence interval. Output drift is the MSE between the decoded outputs from perturbed and clean latents.

\begin{table*}[!t]
  \centering
  \caption{\textbf{Robustness to relative Gaussian perturbations of each method's native latent.} PSNR gains and degradation differences are paired across the same images. The final column reports the reduction in output-drift MSE achieved by RAC relative to TAESD.}
  \label{tab:latent_perturbation}
  \scriptsize
  \setlength{\tabcolsep}{3.5pt}
  \resizebox{\textwidth}{!}{%
  \begin{tabular}{cccccccc}
    \toprule
    Relative $\sigma$ & TAESD PSNR $\uparrow$ & RAC-4 PSNR $\uparrow$ & RAC paired gain $\uparrow$ & TAESD $\Delta$PSNR & RAC-4 $\Delta$PSNR & RAC robustness gain $\uparrow$ & Output-drift reduction $\uparrow$ \\
    \midrule
    0.0 & 24.157 & 24.694 & $+0.537\pm0.060$ & 0.000 & 0.000 & 0.000 & -- \\
    0.1 & 23.675 & 24.355 & $+0.680\pm0.054$ & $-0.482$ & $-0.339$ & $+0.143\pm0.011$ & 80.2\% \\
    0.2 & 22.469 & 23.441 & $+0.972\pm0.047$ & $-1.688$ & $-1.253$ & $+0.435\pm0.026$ & 70.1\% \\
    0.3 & 20.927 & 22.154 & $+1.226\pm0.045$ & $-3.229$ & $-2.540$ & $+0.690\pm0.037$ & 60.8\% \\
    0.5 & 17.812 & 19.141 & $+1.329\pm0.050$ & $-6.344$ & $-5.552$ & $+0.792\pm0.048$ & 44.6\% \\
    \bottomrule
  \end{tabular}}
\end{table*}

At $\sigma=0$, RAC-4 improves PSNR by 0.537 dB. As the latent shift increases, its paired advantage grows to 1.329 dB. More importantly, RAC degrades less at every nonzero perturbation level and reduces output drift by 44.6--80.2\%, showing that multi-step decoding is less sensitive to latent error and can progressively correct deviations along its trajectory.

\subsection{Scaled-Decoder Generation Quality and Efficiency}
\label{sec:latency_cost}

We first evaluate the scaled RAC decoders with the frozen DiT-XL backbone reported by RAE~\cite{zheng2026diffusion}. Every configuration uses the same backbone checkpoint trained for 80 epochs, the same sampling and evaluation protocol, and no classifier-free guidance. The SD-VAE baseline is the single-step result reported by RAE; each RAC variant changes only the decoder under this matched setting.

\begin{wraptable}{r}{0.57\columnwidth}
  \vspace{-8pt}
  \centering
  \caption{\textbf{Scaled RAC generation with frozen DiT-XL.} All settings use 80 epochs without CFG; GFLOPs measure decoder cost.}
  \label{tab:scaled_decoder_generation}
  \scriptsize
  \setlength{\tabcolsep}{1.5pt}
  \resizebox{\linewidth}{!}{%
  \begin{tabular}{lccccc}
    \toprule
    Decoder & gFID$\downarrow$ & IS$\uparrow$ & Prec.$\uparrow$ & Rec.$\uparrow$ & GFLOPs$\downarrow$ \\
    \midrule
    SD-VAE-0.1$\times$ & 6.98 & 126.4 & 0.70 & 0.65 & \textbf{94.3} \\
    SD-VAE-0.2$\times$ & 6.86 & 127.9 & \textbf{0.71} & 0.65 & 152.4 \\
    SD-VAE-0.3$\times$ & \textbf{6.74} & \textbf{129.2} & \textbf{0.71} & \textbf{0.66} & 254.5 \\
    \midrule
    SD-VAE~\cite{zheng2026diffusion} & 7.13 & 124.9 & 0.70 & 0.65 & 310.4 \\
    \bottomrule
  \end{tabular}}
  \vspace{-8pt}
\end{wraptable}

All three scaled RAC decoders improve gFID and IS over the single-step SD-VAE while using fewer decoder GFLOPs. The 0.1$\times$ model reduces gFID from 7.13 to 6.98 with 94.3 GFLOPs, and the 0.3$\times$ model reaches the strongest overall generation result with 6.74 gFID, 129.2 IS, and 0.66 recall. These controlled results directly show that the reconstruction gains from decoder scaling extend to end-to-end generation quality.

We next measure the complete compact-latent-to-image decoding process on one Quadro RTX 8000 at FP16 and $256\times256$ resolution, with 20 warm-up runs followed by 100 CUDA-event timed runs. Table~\ref{tab:decoder_latency} reports decoder-local cost under the default $K=4$ setting.

\begin{table}[H]
  \centering
  \caption{\textbf{Decoder-local efficiency under matched hardware and precision.} Batch-8 latency is reported per image.}
  \label{tab:decoder_latency}
  \scriptsize
  \setlength{\tabcolsep}{3.0pt}
  \resizebox{\columnwidth}{!}{%
  \begin{tabular}{lccccc}
    \toprule
    Method & NFE & Total params. & B1 latency & B8 latency/image & B1 peak memory \\
    \midrule
    Standard SD-VAE & 1 & 83.65M & $29.059\pm0.117$ ms & 31.144 ms & 319.2 MiB \\
    RAC (default) & 4 & 49.51M & $106.421\pm0.284$ ms & 125.659 ms & 319.4 MiB \\
    \bottomrule
  \end{tabular}}
\end{table}

The four-step RAC decoder has $3.66\times$ the Batch-1 latency of the single-step SD-VAE decoder, while reducing total autoencoder parameters from 83.65M to 49.51M (40.82\%) and retaining nearly identical peak activation memory. The integration step count provides explicit control over this quality--latency trade-off.

At the pipeline level, decoder computation remains small. SiT-XL/2 requires approximately 118.6 GFLOPs per forward pass and 29,650 GFLOPs over 250 sampling steps. The lightweight RAC variants require 94.3--254.5 GFLOPs, only 0.32\%--0.86\% of the backbone's sampling cost. Thus, multi-step decoding raises decoder-local latency but adds little end-to-end computation.

\subsection{Additional Experimental Protocol Clarifications}
\label{sec:experimental_protocol_clarifications}

Several evaluated works introduce a coupled system consisting of a generative backbone and its associated autoencoder or decoder. Generation metrics therefore characterize the complete backbone--decoder configuration. Our tables expose individual components in the method labels so that each comparison identifies the axis being varied. Consequently, the same system-level result can appear in more than one table when it is the shared intersection of two analyses. For example, the SD-VAE configuration in Table~\ref{tab:different_vae} uses the default SiT-L backbone and therefore coincides with the SiT-L configuration in Table~\ref{tab:different_size}, which uses SD-VAE. Likewise, the SiT-XL REPA baseline shared by Tables~\ref{tab:different_size} and~\ref{tab:abl} refers to the same 100K-iteration (20-epoch) configuration. These repeated entries preserve a common experimental anchor across VAE, model-scale, and training analyses.

Within every VAE block in Table~\ref{tab:different_vae}, the first row defines the common vanilla system. The rows prefixed by ``+'' are mutually exclusive branches from this shared baseline. The ``+REPA-E'' branch uses the corresponding REPA-E update, while the ``+RAC'' branch uses a single RAC decoder trained with supervision from the frozen baseline encoder. Thus, ``+RAC'' replaces the decoder branch associated with the baseline configuration; the resulting system contains one RAC decoder, with REPA-E and RAC evaluated separately. The host generative backbone, dataset, resolution, training budget, sampling schedule, and evaluation protocol are matched within each comparison. Published baseline entries follow the configurations reported by their originating studies, and RAC is evaluated by changing the designated autoencoder/decoder component under the matched host setup.

This protocol also explains how a frozen teacher encoder is compatible with improved generation metrics. The frozen encoder supplies stable target latents while RAC is trained; the attached SiT, DiT, or UNet supplies latents during generation, and the RAC trajectory maps those latents into images. Metrics such as gFID, sFID, IS, precision, and recall are computed from the final decoded images and therefore reflect the coupled behavior of the generative backbone and decoder. A stronger decoder can improve these image-space metrics while the teacher encoder remains fixed. Equal values in an individual metric across distinct configurations can occur independently; the complete metric vector and the stated backbone--decoder configuration identify each result.


\end{document}